\pdfoutput=1
\documentclass[10pt,twocolumn,twoside,journal,compsoc]{IEEEtran}

\usepackage{cs}
\usepackage{mathsymb}
\usepackage{verbatim}
\usepackage{subfigure}
\usepackage{url}
\usepackage{graphicx}
\usepackage{grffile} %
\usepackage{slashbox,pict2e}
\graphicspath{{./}{./Fig/}{./Figs/}{./Fig/eps/}{./Fig/pdf/}}
\usepackage[nocompress]{cite}
\usepackage{eucal}

\usepackage{algorithm2e}

\let\savedalgorithm\algorithm
\let\savedendalgorithm\endalgorithm
\newenvironment{algorithmic}{%
\savedalgorithm
}{%
\savedendalgorithm
}

\usepackage{algorithm}
\usepackage[pagebackref=true,breaklinks=true,letterpaper=true,colorlinks,bookmarks=false]{hyperref}

\usepackage[footnotesize]{caption}

\newtheorem{prop}{Proposition} %

\usepackage{grffile} %
\usepackage{slashbox,pict2e}
\usepackage{amsmath}
\usepackage{url}
\usepackage{multirow}

\def\Integer{\mathbb{Z}^+}

\renewcommand{\etc}{etc.}

\renewcommand{\paragraph}{\textbf}

\def\cone{{\ding{172}}}
\def\ctwo{{\ding{173}}}
\def\cthree{{\ding{174}}}
\def\cfour{{\ding{175}}}
\def\cfive{{\ding{176}}}

\def\ccone{{\ding{182}}}
\def\cctwo{{\ding{183}}}
\def\ccthree{{\ding{184}}}

\def\ccone{{\ding{172}}}
\def\cctwo{{\ding{173}}}
\def\ccthree{{\ding{174}}}

\def\algoname{{\rm pAUCEns}\xspace}
\def\algonametight{{\rm pAUCEns${\text{{T}}}$}\xspace}

\def\AUC{{\rm{AUC} }}
\def\pAUC{{\rm{pAUC} }}
\def\FPR{{\rm{FPR} }}
\def\nFPPI{{\rm{FPPI}}}

\def\HOGSVM{{{$\tt HOG$}+{$\tt SVM$}}\xspace}
\def\HOGBOOST{{$\tt HOG$}+{$\tt Boosting$}\xspace}

\def\Real{\mathbb{R}}
\def\sC{\mathcal{C}}
\def\sX{\mathcal{X}}
\def\sY{\mathcal{Y}}
\def\sH{\mathcal{H}}

\def\sY{\mathcal{Y}}
\def\sZ{\mathcal{Z}}
\def\bxip{{\bx_i^{+}}}
\def\bxjm{{\bx_j^{-}}}

\def\yij{y_{ij}}

\def\iprime{{l}}

\def\bone{{\boldsymbol 1}}

\def\deltapauc{{ \Delta_{(\alpha,\beta)} }}
\def\tmax{{t_{\mathrm{max}}}}

\def\bpi{{\boldsymbol \pi}}
\def\piij{{\pi_{ij}}}
\def\bpiast{{{\boldsymbol \pi}^{\ast}}}
\def\bPi{{\boldsymbol \Pi}}

\def\phizeta{{\phi_{\zeta}}}

\def\jalpha{{j_{\alpha}}}

\def\jbeta{{j_{\beta}}}

\def\sYmn{{\sY_{m,\jbeta}}}
\def\bYast{{{\boldsymbol Y}^{\ast}}}
\def\bYhat{{\hat{\bY}}}
\def\bYbar{{\bar{\bY}}}

\def\bHp{{\bH_{+}}}
\def\bHm{{\bH_{-}}}

\def\bYast{{\bY^{\ast}}}

\def\vh{{\bh}}

\def\mS{{\bS}}

\def\bxip{\bx_i^{+}}
\def\bxjm{\bx_j^{-}}

\def\bxjfzm{\bx_{(j)_{f|\zeta}}^{-}}
\def\bxkjm{\bx_{k_j}^{-}}
\def\bxjfzm{\bx_{(j)_{f|\zeta}}^{-}}
\def\nbeta{{n_\beta}}
\def\bxjfm{\bx_{(j)_{f}}^{-}}
\def\bxjfmOne{\bx_{(1)_{f}}^{-}}
\def\bxjfmTwo{\bx_{(2)_{f}}^{-}}

\def\bxjfmBeta{\bx_{(\jbeta)_{f}}^{-}}
\def\bxjfzmOne{\bx_{(1)_{f|\zeta}}^{-}}

\def\bxjfzmBeta{\bx_{(\jbeta)_{f|\zeta}}^{-}}
\def\bxjfmN{\bx_{(n)_{f}}^{-}}

\def\UINT128{{\texttt{UINT}$128$}}

\def\bOne{{\bf 1}}

\def\svmpauctight{{\rm SVM$_{ \,\text{pAUC} }^{\,\text{tight} }$}\xspace}

\newcommand{\bigO}{\ensuremath{\mathcal{O}}}

\def\SCOV{{\rm sp-Cov}\xspace}
\def\nSCOV{{\rm sp-Cov}}
\def\SLBP{{\rm sp-LBP}\xspace}
\def\nSLBP{{\rm sp-LBP}}
\def\ChnFtrs{{$\tt ChnFtrs$}\xspace}

\def\ACF{{$\tt ACF$}\xspace}
\def\nACF{{$\tt ACF$}}

\def\atan2{{{\tt atan2}\xspace}}

\def\varnothing{\emptyset}

\def\bpi{{\boldsymbol \pi}}
\def\bPi{{\boldsymbol \Pi}}

\def\bpiast{{{\boldsymbol \pi}^{\ast}}}

\renewcommand{\yij}{\pi_{ij}}
\renewcommand{\bY}{\bpi}
\renewcommand{\sY}{\bPi}

\begin{document}

\title{Pedestrian Detection with Spatially Pooled Features and Structured Ensemble Learning}

\author{Sakrapee Paisitkriangkrai, Chunhua Shen, Anton van den Hengel
    \thanks{
        The authors are with School of Computer Science,
        The University of Adelaide,  SA 5005, Australia.
        C. Shen and A. van den Hengel are also with Australian Research Council Centre of Excellence for
        Robotic Vision.

         Corresponding author: C. Shen
        (e-mail: chunhua.shen@adelaie.edu.au).
    }
}

\IEEEcompsoctitleabstractindextext{
\begin{abstract}

Many typical applications of object detection operate within a
prescribed false-positive range.  In this situation the performance of
a detector should be assessed on the basis of the area under the ROC
curve over that range, rather than over the full curve, as the
performance outside the range is irrelevant.
This measure is labelled as the partial area
under the ROC curve (pAUC).
We propose a novel ensemble learning method which
achieves a maximal detection rate at a user-defined range of false
positive rates by directly optimizing the partial AUC using
structured  learning.
In order to achieve a high object detection performance, we propose a new
approach to extract low-level visual features based on spatial pooling.
Incorporating spatial pooling improves the translational
invariance and thus the robustness of the detection process.
Experimental results on both synthetic and
real-world data sets demonstrate the effectiveness of our approach,
and we show that it is possible to
train state-of-the-art pedestrian detectors using the proposed
structured ensemble learning method with spatially pooled features.
The result is the current best reported performance
on the Caltech-USA pedestrian detection dataset.

\end{abstract}

\begin{IEEEkeywords}
  Pedestrian detection, boosting, ensemble learning, spatial pooling, structured learning.
\end{IEEEkeywords}
}

\maketitle

\section{Introduction}

Pedestrian detection has gained a great deal of attention in the research community
over the past decade.
It is one of several fundamental topics in computer vision.
The task of pedestrian detection is to identify visible pedestrians
in a given image using knowledge gained through analysis of a set of labelled
pedestrian and non-pedestrian exemplars.
Significant progress has been made in the last decade in this area
due to its practical use in many computer vision applications
such as video surveillance, robotics and human computer interaction.
The problem is made difficult by the inevitable variation in target
appearance, lighting and pose, and by occlusion.
In a recent literature survey on pedestrian detection,
the authors evaluated several pedestrian detectors and
concluded that combining multiple features
can significantly boost the performance of pedestrian detection \cite{Dollar2012Pedestrian}.
Hand-crafted low-level visual features have been applied to several computer vision
applications and shown promising results
\cite{Dalal2005HOG,Tuzel2008Pedestrian,Wang2009HOG,Viola2004Robust}.
Inspired by the recent success of spatial pooling on object recognition
and pedestrian detection problems
\cite{Yang2009Linear,Sermanet2013Pedestrian,Park2013Exploring,Wang2013Regionlets},
we propose to perform the spatial pooling operation to create the new feature type
for the task of pedestrian detection.

\begin{figure*}[t]
    \centering
        \includegraphics[width=0.9\textwidth,clip]{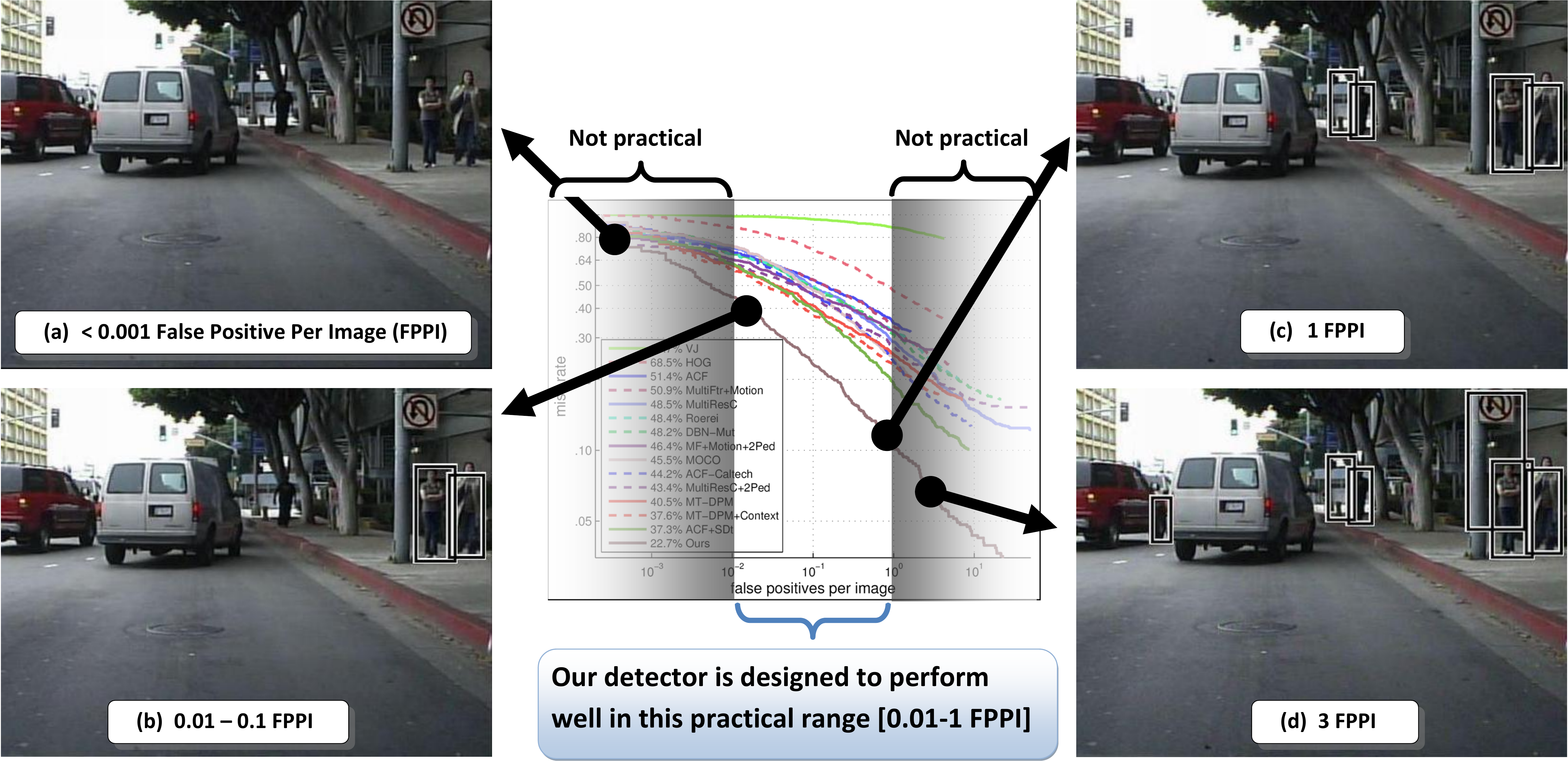}
    \caption{
    An illustration of pedestrian detection performance at four different 
    points on the ROC curve.
    (a) Setting the classification threshold value to be very high 
    such that the detector achieves $< 0.001$ 
    False Positive Per Image (FPPI) is not practical as the detector 
    fails to detect most pedestrians in the image. 
    (b) The detector that achieves a false positive rate 
    between $0.01 - 1$ FPPI is more practical in 
    real-world applications as the detector detects most pedestrians 
    with very few false detections on average. 
    In this example, the detector fails to detect one pedestrian in the image. 
    (c) This example illustrates the pedestrian detector that detects 
    all pedestrians in the image with one false detection on average. 
    (d) By setting the classification threshold value to be very small, 
    we achieve a very high detection rate (the detector detect all 
    pedestrians in the image) at the expense of increased false positives. 
    Existing pedestrian detectors focus their performance    
    in all scenarios, \ie, (a) $-$ (d).
    As a result, the detection performance in the practical range,
    \ie, between scenario (b) and (c), can be
    sub-optimal.
    In contrast, our proposed detector has been designed to perform 
    well only in the region, in which the performance is
    needed for real-world vision applications. 
    }
    \label{fig:illustration}
\end{figure*}

Once the detector has been trained,
the most commonly adopted evaluation method by which to compare the detection performance
of different algorithms is the Receiver Operating Characteristic (ROC) curve.
The curve illustrates the varying performance of a binary classifier
system as its discrimination threshold is altered.
In the face and human detection literature, researchers are often
interested in the low false positive area of the ROC curve since this
region characterizes
the performance needed for most
real-world vision applications (see Fig.~\ref{fig:illustration}).
An algorithm that achieves a high detection rate with many
false positives would be less preferable 
than the algorithm that achieves a moderate detection
rate with very few false positives.
For human detection, researchers often report the partial area
under the ROC curve (pAUC), typically over the range $0.01$ and $1.0$
false positives per image \cite{Dollar2012Pedestrian}.
As the name implies, pAUC is calculated as the
area under the ROC curve between two specified
false positive rates (FPRs).
It summarizes the practical
performance of a detector and often is the
primary performance measure of interest.

Although pAUC is {\em the metric of interest} that has been adopted to
evaluate detection performance,
many classifiers do not
directly optimize this evaluation criterion, and as a result,
often under-perform.
In this paper, we present a principled approach for learning an ensemble classifier
which directly optimizes the \emph{partial} area under the ROC curve,
where the range over which the area is calculated may be selected according to the desired application.
Built upon the structured learning framework,
we thus propose here a novel form of ensemble classifier
which directly optimizes the partial AUC score, which we call \algonametight.
As with all other boosting algorithms, our approach learns a predictor by
building an ensemble of weak classification rules. %
It also relies on a sample re-weighting mechanism to pass the information
between each iteration.
However, unlike traditional boosting, at each iteration,
the proposed approach places a greater emphasis on
samples which have the incorrect ordering\footnote{The positive sample
has an incorrect ordering if it is ranked below the negative sample.
In other words, we want all positive samples to be ranked above all
negative samples.}
to achieve the {\em optimal partial AUC score}.
The result is the ensemble learning method which yields the scoring function
consistent with the correct relative ordering of positive and negative samples
and optimizes the partial AUC score in
a false positive rate range $[\alpha, \beta]$
where $0 \leq \alpha < \beta \leq 1$.

\subsection{Main contributions}
The main contributions of our work can be summarized as follows.

\begin{itemize}
\item We propose a novel approach to extract low-level visual features
based on spatial pooling for the problem of pedestrian detection.
Spatial pooling has been successfully applied in
sparse coding for generic image classification problems.
We show that spatial pooling applied to commonly-used features such as covariance features
\cite{Tuzel2008Pedestrian} and LBP descriptors \cite{Wang2009HOG} improves
accuracy of pedestrian detection.

\item

We propose a new structured  ensemble learning approach which explicitly
optimizes the partial area under the ROC curve (pAUC) between any two given false positive rates.
The method is of particular interest in the wide variety of
applications where performance is most important over a particular
range within the ROC curve. The proposed ensemble learning  is termed \algonametight
(pAUC ENSemble learning with a Tight bound).
The approach shares similarities with conventional boosting methods,
but differs significantly in that the proposed method optimizes
a multivariate performance measure using structured learning.
Our design is simple and a conventional boosting-based visual detector
can be transformed into a \algonametight-based visual detector with
few modifications to the existing code.

Our approach is efficient
since it exploits both the efficient weak classifier training
and the efficient cutting plane solver for optimizing the
partial AUC score in the structured SVM setting.
To our knowledge, our approach is the first principled ensemble method
that directly optimizes the partial AUC in an arbitrary false positive range
$[\alpha, \beta]$.

\item

  We build the {\it best known pedestrian detector} by combining the these two new techniques.
  Experimental results on several data sets, especially on
  challenging human detection data sets, demonstrate the
effectiveness of the proposed approach.
The new approach outperforms all previously reported pedestrian detection results
and achieves state-of-the-art performance on INRIA, ETH,
TUD-Brussels and Caltech-USA pedestrian detection benchmarks.

\end{itemize}

Early versions of our work  \cite{Paul2013Efficient}  introduced
a pAUC-based node classifier for cascade classification,
which optimizes the detection rate in the \FPR range around $[0.49, 0.51]$.
and the low-level visual features based on spatial pooling \cite{Paul2014Strengthening}.
Here we train a single strong classifier with a new structured  learning formulation
which has a tighter convex upper bound on the partial AUC risk compared to \cite{Paul2013Efficient}.
A region proposals generation, known as binarized normed gradients (BING) \cite{Cheng2014Bing},
is applied to speed up the evaluation time of
detector.
We have also introduced new image features and a few careful design when learning the detector.
This leads to a further improvement in accuracy and evaluation time
as compared to \cite{Paul2013Efficient, Paul2014Strengthening}.
Our new detection framework outperforms all reported pedestrian detectors (at the time of submission),
including several complex detectors such as LatSVM \cite{Felzenszwalb2010Object}
(a part-based approach which models unknown parts as latent variables),
ConvNet \cite{Sermanet2013Pedestrian} (deep hierarchical models)
and DBN-Mut \cite{Ouyang2013Modeling}
(discriminative deep model with mutual visibility relationship).

\subsection{Related work}
\label{sec:related}
A few pedestrian detectors have been proposed over the past
decade along with newly created pedestrian detection benchmarks such as
INRIA, ETH, TUD-Brussels, Caltech and Daimler Pedestrian data sets.
We refer  readers to \cite{Dollar2012Pedestrian} for an excellent
review on pedestrian detection frameworks and benchmark data sets.
In this section, we briefly discuss  some relevant work on object detection
and review several recent state-of-the-art pedestrian detectors
that are not covered in \cite{Dollar2012Pedestrian}.

Recent work in the field of object recognition has considered spatial pooling
as one of crucial key components for computer vision system to achieve state-of-the-art
performance on challenging benchmarks, \eg, Pascal VOC, Scene-15, Caltech,
ImageNet
\cite{Wang2010Locality,Chatfield2011Devil,Krizhevsky2012Imagenet,Chatfield2014Return}.
Spatial pooling is a method to extract visual representation based on encoded local features.
In summary, visual features are extracted from a patch representing a small sub-window of an image.
Feature values in each sub-window are spatially pooled and concatenate to form a final feature vector for classification.
Invariant representation is generally obtained by pooling feature vectors over spatially local neighbourhoods.

The use of spatial pooling has long been part of recognition architectures
such as convolutional networks \cite{Lecun1998Gradient,Bengio2009Learning,CVPR15Depth,CVPR15LiuS}.
Spatial pooling (max pooling) is considered as one of critical key ingredients
behind deep convolutional neural networks (CNN)
which achieves the best performance in recent large scale visual recognition tasks.
In CNN, max pooling has been used to reduce the computational complexity
for upper layers and provide a form of translation invariance.
Spatial pooling is general and can be applied to various coding methods,
such as sparse coding, orthogonal matching pursuit and soft threshold \cite{Coates2011Importance}.
Yang \etal propose to compute an image representation based on sparse codes of SIFT features with
multi-scale spatial max pooling \cite{Wang2010Locality}.
They conclude that the new representation significantly outperforms the linear spatial pyramid matching kernel.
Max pooling achieves the best performance in their experiments compared with
square root of mean squared statistics pooling and the mean of absolute values pooling
due to its robustness to local spatial variations.
To further improve the performance of spatial pooling,
Boureau \etal transform the pooling process to be more selective
by applying pooling in both image space and descriptor space \cite{Boureau2011Ask}.
The authors show that this simple technique can significantly boost
the recognition performance even with relatively small dictionaries.

Various ensemble classifiers have been proposed in the literature.
Of these, AdaBoost is one the most well known
as it has achieved tremendous success in computer vision
and machine learning applications.
In object detection, the cost of missing a true target is often
higher than the cost of a false positive.
Classifiers that are optimal under the symmetric cost,
and thus treat false positives and negatives equally,
cannot exploit this information \cite{Viola2002Fast,Human2008Paul}.
Several cost sensitive learning algorithms, where the classifier weights a positive
class more heavily than a negative class, have thus been proposed.

Viola and Jones introduced the asymmetry property
in Asymetric AdaBoost (AsymBoost) \cite{Viola2002Fast}.
However, the authors reported that this asymmetry is immediately
absorbed by the first weak classifier.
Heuristics are then used to avoid this problem.
In addition, one needs to carefully cross-validate this asymmetric parameter
in order to achieve the desired result.
Masnadi-Shirazi and Vasconcelos \cite{Masnadi2011Cost} proposed a
cost-sensitive boosting algorithm based on the
statistical interpretation of boosting.
Their approach is to optimize
the cost-sensitive loss by means of gradient descent.
Most work along this line addresses the pAUC evaluation criterion {\em
indirectly}. In addition, one needs to carefully cross-validate the asymmetric
parameter in order to maximize the detection rate in a particular false
positive range.

Several algorithms that directly optimize the pAUC score
have been proposed in bioinformatics
\cite{Hsu2012Linear,Komori2010Boosting}.
Dodd and Pepe propose a regression modeling framework based on the pAUC score \cite{Pepe2000Combining}.
Komori and Eguchi  optimize the pAUC
using boosting-based algorithms \cite{Komori2010Boosting}.
Their algorithm is heuristic in nature.
Narasimhan and Agarwal develop  structural SVM based methods which directly optimize
the pAUC score \cite{Narasimhan2013Structural,Narasimhan2013SVM}.
They demonstrate that their approaches significantly outperform several existing algorithms,
including pAUCBoost \cite{Komori2010Boosting} and asymmetric SVM
\cite{Wu2008Asymmetric}.
Building on Narasimhan and Agarwal's work,
we propose the principled fully-corrective ensemble method which
directly optimizes the pAUC evaluation criterion.
The approach is flexible and can be applied to an arbitrary false
positive range $[\alpha, \beta]$.
To our knowledge, our approach is the first principled ensemble learning method
that directly optimizes the partial AUC in a false positive range
not bounded by zero.
It is important to emphasize here the difference between our approach
and that of \cite{Narasimhan2013Structural}.
In \cite{Narasimhan2013Structural} the authors train a linear structural SVM
while our approach learns the ensemble of classifiers.
A few recently proposed pedestrian detectors are as follows.
Sermanet \etal train a pedestrian detector using a
convolutional network model \cite{Sermanet2013Pedestrian}.
Instead of using hand designed features, they propose to
use unsupervised sparse auto encoders to automatically learn features
in a hierarchy.
The features generated from a multi-scale convolutional network capture
both global and local details such as shapes, silhouette and facial components.
Experimental results show that their detector achieves competitive results
on major benchmark data sets.
Benenson \etal investigate different low-level aspects of
pedestrian detection \cite{Benenson2013Seeking}.
The authors show that by properly tuning low-level features,
such as
feature selection, pre-processing the raw image
and classifier training,
it is possible to reach state-of-the-art results on major benchmarks.
From their paper, one key observation that significantly improves
the detection performance is to
apply image normalization to the test image before extracting features.
Park \etal propose new motion features for detecting pedestrians
in a video sequence \cite{Park2013Exploring}.
The authors use optical flow to align large objects in a sequence
of image frames and use temporal difference features to capture the
information that remains.
By factoring out camera motion and combining
their proposed motion features with channel features \cite{Dollar2009Integral},
the new detector achieves a five-fold reduction in false positives
over previous best results on the Caltech pedestrian benchmark.

Another related work that applies structured SVM learning to
object detection is the work of Desai \etal \cite{Desai2011Discriminative}.
The authors train the model that captures
the spatial arrangements of various object classes in the image
by considering which spatial layouts of objects to suppress 
and which spatial layouts of objects to favor.
Although both our approach and \cite{Desai2011Discriminative} cast the problem
as a structured prediction and apply the cutting plane optimization,
the underlying assumptions and resulting models are quite different.
The formulation of \cite{Desai2011Discriminative} 
focuses on incorporating geometric configurations
between multiple object classes instead of 
optimizing the detection
rate within a prescribed false positive range.
Our approach learns a function 
that optimizes the partial AUC risk between two false positive rates.
In addition, it is {\em not trivial} to extend the formulation of
\cite{Desai2011Discriminative} to boosting setting.

\subsection{Notation}
Vectors are denoted by lower-case bold letters, \eg, $ \bx $,
matrices are denoted by upper-case bold letters, \eg, $\bX$
and sets are denoted by calligraphic upper-case letters, \eg, $\sX$.
All vectors are assumed to be column vectors.
The $ ( i, j ) $ entry of $ \bX$ %
is $x_{ij}$.
Let $\{ \bxip \}_{i=1}^m$ be a set of pedestrian training examples,
$\{ \bxjm \}_{j=1}^n$ be a set of non-pedestrian training examples
and $\bx \in \Real^{d}$ be a $d$ dimensional feature vector.
The tuple of all training samples is written as
$\mS = (\mS_{+}, \mS_{-})$ where
$\mS_{+} = (\bx_1^{+}, \cdots, \bx_m^{+}) \in \sX^m$ and
$\mS_{-} = (\bx_1^{-} , \cdots, \bx_n^{-}) \in \sX^n$.
We denote by $\sH$ a set of all possible outputs of weak learners.
Assuming that we have $\tau$ possible weak learners,
the output of weak learners for positive and negative data
can be represented as $\bH = (\bHp, \bHm)$   %
where $\bHp \in \Real^{\tau \times m}$
and  $\bHm \in \Real^{\tau \times n}$, respectively.
Here $h^{+}_{ti}$ is the label predicted by the weak learner
$\hbar_t(\cdot)$ on the positive training data $\bxip$.
Each column $\vh_{:l}$ of the matrix $\bH$ represents the output of all weak learners
when applied to the training instance $\bx_{l}$.
Each row $\vh_{t:}$ of the matrix $\bH$ represents the output predicted
by the weak learner $\hbar_t(\cdot)$ on all the training data.
In this paper, we are interested in the partial \AUC
(area under the ROC curve)
within a specific
false positive range $[\alpha, \beta]$.
Given $n$ negative training samples, we let $\jalpha = \lceil n \alpha \rceil$
and $\jbeta =  \lfloor n \beta \rfloor$.
Let $\sZ_{\beta} = \bigl(\begin{smallmatrix} \mS_{-} \\ \jbeta \end{smallmatrix} \bigr)$
denote the
set of all subsets of negative training instances of size $\jbeta$.
We define $\zeta = \{ \bxkjm \}_{j=1}^{\jbeta} \in \sZ_{\beta}$ as a given
subset of negative instances,
where $\boldsymbol{k} = [k_1,\ldots ,k_{\jbeta}]$ is a vector indicating which elements of $\mS_{-}$ are included.
The goal is to learn a set of binary weak learners and a scoring function,
$f: \Real^{d} \rightarrow \Real$, 
that acheive good performance in terms of the pAUC between
some specified false positive rates $\alpha$ and $\beta$
where $0 \leq \alpha < \beta \leq 1$.
Here
$f(\bx) = \sum_{t=1}^\tau w_{t} \hbar_t(\bx)$
where $\bw \in \Real^{\tau}$ is the linear coefficient vector,
$\{\hbar_t(\cdot)\}_{t=1}^{\tau}$ denote a set of binary weak 
learners and $\tau$ is the number of weak learners.

\section{Our approach}
\label{sec:approach}

Despite several important work on object detection, the most practical and successful
pedestrian detector is still the sliding-window based method of Viola and Jones \cite{Viola2004Robust}.
Their method consists of two main components: feature extraction and the AdaBoost classifier.
For pedestrian detection, the most commonly used features
are HOG \cite{Dalal2005HOG} and HOG+LBP \cite{Wang2009HOG}.
Doll\'ar \etal propose Aggregated Channel Features (\nACF) which combine
gradient histogram (a variant of HOG), gradients and LUV \cite{Dollar2014Fast}.
ACF uses the same channel features as \ChnFtrs \cite{Dollar2009Integral},
which is shown to outperform HOG \cite{Dollar2009Integral,Benenson2013Seeking}.

To train the classifier, the procedure known as bootstrapping is often applied, which
harvests hard negative examples and re-trains the classifier.
Bootstrapping can be repeated several times.
It is shown in \cite{Walk2010New} that at least two bootstrapping iterations
are required for the classifier to achieve good performance.
In this paper, we design the new pedestrian detection framework based
on the new spatially pooled features,
a novel form of ensemble classifier which directly optimizes the partial area
under the ROC curve and
an efficient region proposals generation.
We first propose the new feature type based on a modified low-level descriptor
and spatial pooling.
In the next section, we discuss how the performance measure can be further
improved using the proposed structured learning framework.
Finally, we discuss our modifications to \cite{Dollar2014Fast} in order to achieve
state-of-the-art detection results on Caltech pedestrian detection benchmark data sets.

\subsection{Spatially pooled features}

Spatial pooling has been proven to be invariant to various image transformations and
demonstrate better robustness to noise \cite{Boureau2011Ask,Chatfield2011Devil,Coates2011Importance}.
Several empirical results have indicated that a pooling operation can greatly improve the recognition performance.
Pooling combines several visual descriptors obtained at nearby locations into some statistics that
better summarize the features over some region of interest (pooling region).
The new feature representation preserves visual information over a local neighbourhood
while discarding irrelevant details and noises.
Combining max-pooling with unsupervised feature learning methods have led to
state-of-the art image recognition performance on several object recognition tasks.
Although these feature learning methods have shown promising results over hand-crafted features,
computing these features from learned dictionaries is still a time-consuming process
for many real-time applications.
In this section, we further improve the performance of low-level features by adopting
the pooling operator commonly applied in unsupervised feature learning.
This simple operation can enhance the feature robustness to noise and image transformation.
In the following section, we investigate two visual descriptors which have shown to complement HOG
in pedestrian detection, namely covariance descriptors and LBP.
It is important to point out here that our approach is not limited to these two features,
but can be applied to any low-level visual features.

\paragraph{Background}
A covariance matrix is positive semi-definite.
 It provides a measure of the relationship between two or more sets of variates.
The diagonal entries of covariance matrices represent the variance of each feature and
the non-diagonal entries represent the correlation between features.
The variance measures the deviation of low-level features from the mean and provides
information related to the distribution of low-level features.
The correlation provides the relationship between multiple low-level features within the region.
In this paper,
we follow the feature representation as proposed in \cite{Tuzel2008Pedestrian}.
However, we introduce an additional edge orientation which considers
the sign of intensity derivatives.
Low-level features used in this paper are:
\begin{align}
  \notag
\left[  x, \; y, \; |I_x|, \; |I_y|, \; |I_{xx}|, \; |I_{yy}|, \; M, \; O_1, \; O_2 \right]
\end{align}
where $x$ and $y$ represent the pixel location, and
$I_x$ and $I_{xx}$ are first and second intensity derivatives along the $x$-axis.
The last three terms are the gradient magnitude
($M = \sqrt{I_x^2 +I_y^2}$),
edge orientation as in \cite{Tuzel2008Pedestrian} ($O_1 = \arctan ( |I_y| / | I_x| )$)
and an additional edge orientation $O_2$ in which,
\begin{align}
  \notag
O_2 =
            \begin{cases}
                \atan2(I_y,I_x)  \quad& \text{if} \; \atan2(I_y,I_x) > 0, \\
                \atan2(I_y,I_x) + \pi  \quad& \text{otherwise.}
            \end{cases}
\end{align}
The orientation $O_2$ is mapped over the interval $[0,\pi]$.
Although some $O_1$ features might be redundant after introducing $O_2$,
these features would not deteriorate the performance as they
will not be selected by the weak learner.
Our preliminary experiments show that using $O_1$ alone yields
slightly worse performance than combining $O_1$ and $O_2$.
With the defined mapping, the input image is mapped to a $9$-dimensional feature image.
The covariance descriptor of a region is a $9 \times 9$ matrix, and due to symmetry,
only the upper triangular part is stored,
which has only $45$ different values.

Local Binary Pattern (LBP) is a texture descriptor that represents the binary code of each image patch
into a feature histogram \cite{Ojala2002Multi}.
The standard version of LBP is formed by thresholding the $3 \times 3$-neighbourhood of each pixel with the centre pixel's value.
All binary results are combined to form an $8$-bit binary value ($2^8$ different labels).
The histogram of these $256$ different labels can be used as texture descriptor.
The LBP descriptor has shown to achieve good performance in many texture classification \cite{Ojala2002Multi}.
In this work, we adopt an extension of LBP, known as the uniform LBP, which can better filter out noises \cite{Wang2009HOG}.
The uniform LBP is defined as the binary pattern that contains at most two bitwise transitions from $0$ to $1$ or vice versa.

\begin{figure}[t]
    \centering
        \includegraphics[width=0.4\textwidth,clip]{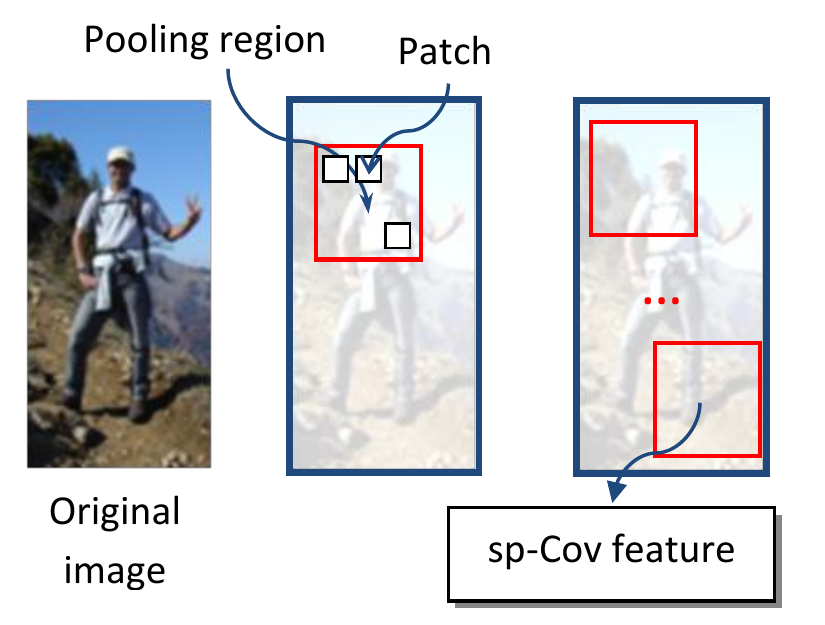}
    \caption{
    Architecture of our pooled features.
    In this example, \SCOV are extracted from each fixed sized pooling region.
    }
    \label{fig:illus}
\end{figure}

\paragraph{Spatially pooled covariance}
In this section, we improve the spatial invariance and robustness of the
original covariance descriptor by applying the operator known as spatial pooling.
There exist two common pooling strategies in the literature:
average pooling and max-pooling.
We use max-pooling as it has been shown to outperform average
pooling in image classification \cite{Coates2011Importance,Boureau2011Ask}.
We divide the image window into {\em small patches}
(refer to Fig.~\ref{fig:illus}).
For each patch, covariance features are calculated over pixels within
the patch.
For better invariance to translation and deformation,
we perform spatial pooling over a pre-defined spatial region
({\em pooling region}) and use the obtained results
to represent covariance features
in the pooling region.
The pooling operator thus summarizes multiple covariance matrices
within each pooling region into a
single matrix which represents covariance information.
We refer to the feature extracted from each pooling region
as spatially pooled covariance (\nSCOV) feature.
Note that extracting covariance features in each patch
can be computed efficiently using the integral image trick \cite{Tuzel2006Region}.
Our \SCOV differs from covariance features in
\cite{Tuzel2008Pedestrian} in the following aspects:

1. We apply spatial pooling to a set of covariance descriptors in the pooling region.
  To achieve this, we ignore the geometry of covariance matrix
  and stack the upper triangular part of the covariance matrix into a vector
  such that pooling is carried out on the vector space.
  For simplicity, we carry out pooling over
  a square image region of fixed resolution.
  Considering pooling over a set of arbitrary rectangular regions as in \cite{Jia2012Beyond}
  is likely to further improve the performance of our features.

2. Instead of normalizing the covariance descriptor of each patch based on the whole
  detection window \cite{Tuzel2008Pedestrian},
  we calculate the correlation coefficient within each patch.
  The correlation coefficient returns the value in the range $[-1,1]$.
  As each patch is now independent, the feature extraction can be done
  in parallel on the GPU.

\paragraph{Implementation}
We extract \SCOV using multi-scale patches with the following sizes:
$8 \times 8$, $16 \times 16$ and $32 \times 32$ pixels.
Each scale will generate a different set of visual descriptors.
Multi-scale patches have also been used in \cite{Bo2013Multipath}.
In this paper, the use of multi-scale patches is important as it expands
the richness of our feature representations and enables us to capture
human body parts at different scales.
In our experiments, we set the patch spacing stride (step-size)
to be $1$ pixel.
The pooling region is set to be  $4 \times 4$-pixel
and the pooling spacing stride is set to $4$ pixels in our experiments.

\paragraph{Spatially pooled LBP}
Similar to \SCOV, we divide the image window into small patches and extract LBP over pixels within the patch.
The histogram, which represents the frequency of each pattern occurring, is computed over the patch.
For better invariance to translation, we perform spatial pooling over a pooling region and use the obtained results
to represent the LBP histogram in the pooling region.
We refer to the new feature as spatially pooled LBP (\nSLBP) feature.

\paragraph{Implementation}
For the LBP operator, we use the $3 \times 3$-neighbourhood of each pixel and extract the local histogram using
a patch size of $4 \times 4$, $8 \times 8$ and $16 \times 16$ pixels.
For \SLBP, the patch spacing stride, the pooling region and the pooling spacing stride are set to $1$ pixel, $8 \times 8$-pixel and
$4$ pixels, respectively.

\paragraph{Discussion}
Although we make use of spatial pooling, our approach
differs significantly from the unsupervised feature learning pipeline,
which has been successfully
applied to image classification problem \cite{Yang2009Linear,Bo2013Multipath}.
Instead of pooling encoded features over a pre-trained dictionary,
we compute \SCOV and \SLBP by performing pooling directly on
covariance and LBP features extracted from local patches.
In other words, our proposed approach removes the dictionary learning and
feature encoding from the conventional unsupervised feature learning
\cite{Yang2009Linear,Bo2013Multipath}.
The advantage of our approach over conventional feature learning
is that our features have much less dimensions than
the size of visual words often used in generic image classification \cite{Yang2009Linear}.
Using too few visual words can significantly degrade
the recognition performance as reported in \cite{Chatfield2011Devil} and using too many visual words would
lead to very high-dimensional features and thus make the classifier training become computationally infeasible.

\subsection{Optimizing partial AUC}

\paragraph{Structured learning approach}
Before we propose our approach, we briefly review the concept of
\svmpauctight $[\alpha,\beta]$ \cite{Narasimhan2013SVM},
in which our ensemble learning approach is built upon.
The area under the empirical ROC curve (AUC) can be defined as,
\begin{align}
    \label{EQ:auc}
        \AUC = \frac{1}{mn} \sum_{i=1}^m \sum_{j=1}^n \bOne \bigl( f(\bxip) > f(\bxjm) \bigr).
\end{align}
The objective is to learn a scoring function $f$,
$f: \Real^{d} \rightarrow \Real$,
that maximizes the AUC, or equivalently,
minimizes the empirical risk,
\begin{align}
    \label{EQ:auc2}
        R_{\AUC}(f) = 1 - \AUC.
\end{align}
For the partial AUC (pAUC) in the false positive range $\left[ \alpha, \beta \right]$,
the empirical pAUC risk can be written as \cite{Narasimhan2013SVM}:
\begin{align}
    \label{EQ:pauc}
        R_\pAUC(f) = \frac{1}{c} \;
                {\textstyle \sum}_{i=1}^m
                {\textstyle \sum}_{j=\jalpha+1}^{\jbeta} \bone ( f(\bxip) < f(\bxjfzm) ).
\end{align}
Here 
$c$ is a constant, $c = mn(\beta-\alpha)$,
$\bxip$ denotes the $i$-th positive training instance,
$\bxjfzm$ denotes the $j$-th negative training instance sorted by the scoring function, $f$, in
the set
$\zeta = \{ \bxjfzm \}_{j=1}^{\jbeta} \in \sZ_{\beta}$,
$\zeta$
denote the chosen subset of negative instances and
$\sZ_{\beta} = \bigl(\begin{smallmatrix} \mS_{-} \\ \jbeta \end{smallmatrix} \bigr)$
denote the
set of all subsets of negative training instances of size $\jbeta$.
In other words, we sort all negative instances based on
their scoring values to obtain
$\{ \bxjfm \}_{j=1}^{n}$
in which $f(\bxjfmOne)$
$\geq$ $f(\bxjfmTwo)$
$\geq \ldots$ 
$\geq$ $f(\bxjfmBeta)$
$\geq \ldots $
$\geq$ $f(\bxjfmN)$
and $\zeta$ $=$ $\{\bxjfzmOne,\cdots,\bxjfzmBeta\}$.
Although the number of elements in $\zeta$ is $\jbeta$
(there are $\jbeta$ negative samples in the set),
the empirical pAUC risk defined in \eqref{EQ:pauc}
is computed from $\jbeta-\jalpha$ negative samples.

Clearly \eqref{EQ:pauc} is minimal when all positive samples, $\{ \bxip \}_{i=1}^m$,
are ranked above
$\{ \bxjfzm \}_{j=\jalpha+1}^{\jbeta}$,
which represent negative samples
in our prescribed false positive range $[\alpha,\beta]$
(in this case, the log-average miss rate would be zero).
The structural SVM framework can be adopted to optimize the \pAUC risk by
considering a classification problem of all $m \times \jbeta$ pairs of
positive and negative samples.
We define a new label matrix $\bY \in \sY_{m,\jbeta} = \{0, 1\}^{m \times \jbeta}$
(on the entire positive instances $\{ \bxip \}_{i=1}^m$
and a given subset of negative instances
$\zeta = \{ \bxkjm \}_{j=1}^{\jbeta} \in \sZ_{\beta}$
where $\boldsymbol{k} = [k_1,\ldots ,k_{\jbeta}]$ is a vector indicating which elements of $\mS_{-}$ are included)
whose value for the pair $(i,j)$ is defined as:
\begin{align}
    \label{EQ:EQ1}
        \yij  =
            \begin{cases}
                0  \quad& \text{if} \; \bxip \; \text{is ranked above} \; \bxjm  \\
                1  \quad& \text{otherwise.}
            \end{cases}
\end{align}
The true pair-wise label is defined as
$\bYast$ where $\pi_{ij}^{\ast} = 0$ for all pairs $(i,j)$.
The \pAUC loss is calculated from the number of pairs of examples that are ranked in the wrong order, \ie, negative examples are ranked before positive examples. 
Hence the \pAUC loss between the prediction $\bY$ and the true pair-wise label $\bYast$ can be written as:
\begin{align}
    \label{EQ:deltapauc}
        \deltapauc (\bY, \bYast) &=
            \frac{1}{c} \; {\textstyle \sum}_{i=1}^m
                    {\textstyle \sum}_{j=\jalpha+1}^{\jbeta} 
                    \left( \pi_{i,(j)_\bpi} - \pi^{\ast}_{i,(j)_\bpi} \right) \notag \\     
 							  \; &=
            \frac{1}{c} \; {\textstyle \sum}_{i=1}^m
                    {\textstyle \sum}_{j=\jalpha+1}^{\jbeta} 
                    \left( \pi_{i,(j)_\bpi} - 0 \right) \notag \\     
                               \; &=
            \frac{1}{c} \; {\textstyle \sum}_{i=1}^m
                    {\textstyle \sum}_{j=\jalpha+1}^{\jbeta} \pi_{i,(j)_\bpi},
\end{align}
where $(j)_\bpi$ denotes the index of the negative instance in $\bS_{-}$
ranked in the $j$-th position by any fixed ordering consistent with the
matrix
$\bY$.
We define a joint feature map,
$\phizeta:(\sX^m \times \sX^n) \times \sY_{m,\jbeta} \rightarrow \Real^d$,
which takes a set of training instances
($m$ positive samples and $n$ negative samples) and an ordering
matrix of dimension $m \times \jbeta$ and produce a vector output in $\Real^d$ as:
\begin{align}
    \label{EQ:featmap}
        \phizeta (\bS, \bY) =  \frac{1}{c} \; {\textstyle \sum}_{i=1}^m
                   {\textstyle \sum}_{j=1}^{\jbeta} (1-\yij) (\bxip - \bxkjm).
\end{align}
This feature map ensures that the variable $\bw$ ($\bw \in \Real^d$) that optimizes $\bw^\T \phizeta (\bS, \bY)$
will also produce the optimal \pAUC score for $\bw^\T \bx$.
We can summarize the above problem as the following convex optimization
problem \cite{Narasimhan2013SVM}:
\begin{align}
\label{EQ:hinge1}
    \min_{ \bw , \xi }   \quad
    &
    \quad
    \frac{1}{2} \| \bw  \|_{2}^{2} + \nu \, \xi  \quad   \\ \notag
    \st \; &
    \bw^\T ( \phizeta (\bS, \bYast) - \phizeta (\bS, \bY) )
    \geq \deltapauc (\bY, \bYast) - \xi,
\end{align}
$\forall \zeta \in \sZ_{\beta}$,
$\forall \bY \in \sY_{m,\jbeta}$ and $\xi \geq 0$.
Note that $\bYast$ denotes the correct relative ordering and
$\bY$ denotes any arbitrary orderings and $\nu$ controls the amount
of regularization.

\paragraph{Partial AUC based ensemble classifier}
In order to design an ensemble-like algorithm for the pAUC,
we first introduce a projection function,
$\hbar(\cdot)$, which projects an instance vector
$\bx$ to $\{-1,+1\}$.
This projection function is also known as
the weak learner in boosting.
In contrast to the previously described structured learning,
we learn the scoring function, which optimizes the area
under the curve between two false positive rates
of the form:
$f(\bx) = \sum_{t=1}^\tau w_{t} \hbar_t(\bx)$
where $\bw \in \Real^{\tau}$ is the linear coefficient vector,
$\{\hbar_t(\cdot)\}_{t=1}^{\tau}$ denote a set of binary weak 
learners and $\tau$ is the number of weak learners.
Let us assume that we have already learned a set of all projection functions.
By using the same pAUC loss, $\deltapauc(\cdot, \cdot)$, as in \eqref{EQ:deltapauc},
and the same feature mapping, $\phizeta (\cdot, \cdot)$, as in \eqref{EQ:featmap},
the optimization problem we want to solve is:
\begin{align}
\label{EQ:hinge1a}
    \min_{ \bw , \xi }   \quad
    &
    \frac{1}{2} \| \bw  \|_{2}^{2} + \nu \, \xi    \\ \notag
    \st \; &
    \bw^\T ( \phizeta (\bH, \bYast) - \phizeta (\bH, \bY) )
    \geq \deltapauc (\bY, \bYast) - \xi,
\end{align}
$\forall \bY \in \sY_{m,\jbeta}$ and $\xi \geq 0$.
$\bH = (\bHp, \bHm)$ is the projected output
for positive and negative training samples.
$\phizeta (\bH, \bY) = [\phizeta (\bh_{1:}, \bY), \cdots, \phizeta (\bh_{\tau:}, \bY)]$
where $\phizeta (\bh_{t:}, \bY): (\Real^{m} \times \Real^{n}) \times \sY_{m,\jbeta} \rightarrow \Real$
and it is defined as,
\begin{align}
    \label{EQ:featmap2}
    \phizeta (\bh_{t:}, \bY) = \frac{1}{c} \;
           {\textstyle \sum}_{i=1}^m {\textstyle \sum}_{j=1}^{\jbeta} (1 &- \yij) \\ \notag
            &\bigl( \hbar_{t} (\bxip) - \hbar_{t} (\bxkjm) \bigr),
\end{align}
where $\{ \bxkjm \}_{j=1}^{\jbeta}$ is any given subsets of negative instances
and $\boldsymbol{k} = [k_1,\ldots ,k_{\jbeta}]$ is a vector indicating which elements of $\mS_{-}$ are included.
The only difference between \eqref{EQ:hinge1} and \eqref{EQ:hinge1a}
is that the original data is now projected to a new non-linear feature space.
The dual problem of \eqref{EQ:hinge1a} can be written as,
\begin{align}
    \label{EQ:hinge2}
        \max_{ \boldsymbol \lambda }   \quad &
        {\textstyle \sum_{\bY}} \blambda_{(\bY)} \deltapauc(\bYast, \bY) - \\ \notag
        &\quad \frac{1}{2}  {\textstyle \sum_{\bY,\bYhat}}
            \blambda_{(\bY)} \blambda_{(\bYhat)}
            \langle   \phi_{\Delta}(\bH, \bY), \phi_{\Delta}(\bH, \bYhat) \rangle  \\ \notag
        \st \quad &
        0 \leq
              {\textstyle \sum_{\bY}} \blambda_{ (\bY) }
          \leq \nu.
\end{align}
where $\boldsymbol \lambda$ is the dual variable,
$\blambda_{(\bY)}$ denotes the dual variable
associated with the inequality constraint for $\bY \in \sYmn$
and $\phi_{\Delta}(\bH, \bY) = \phizeta (\bH, \bYast)  - \phizeta (\bH, \bY)$.
To derive the Lagrange dual problem, the following KKT condition
is used,
\begin{align}
    \label{EQ:KKT}
        \bw = \sum_{\bY \in \sY_{m,\jbeta}} \blambda_{(\bY)}
        \bigl( \phizeta (\bH, \bYast)  - \phizeta (\bH, \bY) \bigr).
\end{align}

\paragraph{Finding best weak learners}
In this section, we show how one can explicitly learn the projection
function, $\hbar(\cdot)$.
We use the idea of column generation to derive an ensemble-like
algorithm similar to LPBoost \cite{demiriz2002}.
The condition for applying the column generation is that the duality gap
between the primal and dual problem is zero (strong duality).
By inspecting the KKT condition,
at optimality, \eqref{EQ:KKT} must hold for all $t = 1, \cdots, \tau$.
In other words, $w_t = \sum_{\bY \in \sYmn} \blambda_{(\bY)}
\bigl( \phizeta (\bh_{t:}, \bYast)  - \phizeta (\bh_{t:}, \bY) \bigr)$
must hold for all $t$.

For weak learners in the current working set, the corresponding
condition in \eqref{EQ:KKT} is satisfied by the current solution.
For weak learners that are not yet selected,
they do not appear in the current restricted optimization problem
and their corresponding coefficients are zero ($w_t = 0$).
It is easy to see that if $\sum_{\bY \in \sYmn} \blambda_{(\bY)}
\bigl( \phizeta (\bh_{t:}, \bYast)  - \phizeta (\bh_{t:}, \bY) \bigr) = 0$
for all $\hbar_t(\cdot)$ that are not in the
current working set, then the current solution
is already the globally optimal one.
Hence the subproblem for selecting the best
weak learner is:
\begin{align}
    \label{EQ:weak1}
    \hbar^{\ast}(\cdot) = \argmax_{\hbar \in \sH }
    \;
    \Bigl| {\textstyle \sum}_{\bY} \blambda_{(\bY)}
        \bigl(  \phizeta (\bh, \bYast) - \phizeta (\bh, \bY) \bigr) \Bigr|.
\end{align}
In other words, we pick the weak learner with the
value $ | {\textstyle \sum}_{\bY} \blambda_{(\bY)}
        \bigl(  \phizeta(\bh, \bYast) - \phizeta (\bh, \bY) \bigr) |$
most deviated from zero.
Thus, a stopping condition for our algorithm is
\begin{equation}
| {\textstyle \sum}_{\bY} \blambda_{(\bY)}
        \bigl(  \phizeta(\bh, \bYast) - \phizeta (\bh, \bY) \bigr) | < \varepsilon,
        \label{EQ:Stop}
      \end{equation}
      where $ \varepsilon > 0 $ is a small precision constant (\eg, $ 10^{-4}$).
To find the most optimal weak learner in $\sH$,
we consider the relative ordering of all positive and negative instances,
$\bY \in \sY_{m,\jbeta} = \{0, 1\}^{m \times \jbeta}$,
whose value for the pair $(i,j)$ is similar to \eqref{EQ:EQ1}.
Given the weak learner $\hbar(\cdot)$, we define the joint feature
map for the output of the weak learner $\hbar(\cdot)$ as,
\begin{align}
    \label{EQ:featmap_old}
    \phizeta (\bh, \bY) = \frac{1}{c} \;
           {\textstyle \sum}_{i=1}^m {\textstyle \sum}_{j=1}^{\jbeta} (1 - \piij)
            \bigl( \hbar (\bxip) - \hbar (\bxkjm) \bigr).
\end{align}
The subproblem for generating the optimal weak learner at iteration $t$
considering the relative ordering of all positive and negative training instances,
\ie \eqref{EQ:weak1}, can be re-written as,
\begin{align}
    \label{EQ:weak2}
    \notag
    \hbar_t^{\ast}(\cdot) &= \argmax_{\hbar \in \sH }
    \;
    \Bigl| {\textstyle \sum}_{\bpi} \blambda_{(\bpi)}
        \bigl(  \phizeta (\bh, \bpiast) - \phizeta (\bh, \bpi) \bigr) \Bigr| \\ \notag
    &= \argmax_{\hbar \in \sH}
    \;
    \Bigl| \sum_{\bpi} \blambda_{(\bpi)}
        \sum_{i,j} \piij \bigl( \hbar(\bxip) - \hbar(\bxjm) \bigr) \Bigr| \\ \notag
    &= \argmax_{\hbar \in \sH}
    \;
    \Bigl| \sum_{i,j} \bigl( {\textstyle \sum}_{\bpi} \blambda_{(\bpi)} \piij \bigr)
          \bigl( \hbar(\bxip) - \hbar(\bxjm) \bigr) \Bigr| \\ \notag
    &= \argmax_{\hbar \in \sH}
    \;
    \Bigl| {\textstyle \sum}_{ \iprime } u_{\iprime} y_{\iprime} \hbar (\bx_{\iprime}) \Bigr| \\
    &= \argmax_{\hbar \in \sH}
    \;
    {\textstyle \sum}_{ \iprime } u_{\iprime} y_{\iprime} \hbar (\bx_{\iprime})
\end{align}
where $i$, $j$, $\iprime$ index the positive training samples ($i = 1, \cdots, m$),
the negative training samples ($j = 1, \cdots, \jbeta$) and
the entire training samples ($\iprime = 1, 2$,$\cdots$,$m+\jbeta$), respectively.
Here $y_{\iprime}$ is equal to $+1$ if $\bx_{\iprime}$ is a positive sample and
$-1$ otherwise, and
\begin{align}
    \label{EQ:updatew}
        u_{\iprime}  =
            \begin{cases}
                \sum_{\bpi,j} \blambda_{(\bpi)} \pi_{\iprime j}
                    \quad& \text{if} \; \bx_{\iprime} \; \text{is a positive sample}  \\
                \sum_{\bpi,i} \blambda_{(\bpi)} \pi_{i \iprime}
                    \quad& \text{otherwise}.
            \end{cases}
\end{align}
For decision stumps and decision trees, the last equation in \eqref{EQ:weak2} is always valid since
the weak learner set $\sH$ is negation-closed \cite{Komori2011Boosting}.
In other words, if $\hbar(\cdot) \in \sH$, then
$[-\hbar](\cdot) \in \sH$, and vice versa.
Here $[-\hbar](\cdot) = -\hbar(\cdot)$.
For decision stumps, one can flip the inequality sign such that
$\hbar(\cdot) \in \sH$ and $[-\hbar](\cdot) \in \sH$.
In fact, any linear classifiers of the form
$\sign(\sum_{t} a_t x_t + a_{0})$ are negation-closed.
Using \eqref{EQ:weak2} to choose the best weak learner is
not heuristic as the solution to \eqref{EQ:weak1} decreases
the duality gap the most for the current solution.

\paragraph{Optimizing weak learners' coefficients}
We solve for the optimal $\bw$ that minimizes our
objective function \eqref{EQ:hinge1a}.
However, the optimization problem \eqref{EQ:hinge1a} has
an exponential number of constraints, one for each matrix
$\bY \in \sYmn$.
As in \cite{Narasimhan2013Structural, Joachims2009Cutting},
we use the cutting plane method to solve this problem.
The basic idea of the cutting plane is that a small subset
of the constraints are sufficient to find an $\epsilon$-approximate
solution to the original problem.
The algorithm starts with an empty constraint set and it adds the most
violated constraint set at each iteration.
The QP problem is solved using linear SVM and the process
continues until no constraint is violated by more than $\epsilon$.
Since, the quadratic program is of constant size and
the cutting plane method converges in a constant number of iterations,
the major bottleneck lies in the combinatorial optimization (over
$\sYmn$) associated with finding the most violated constraint set
at each iteration.
Narasimhan and Agarwal show how this combinatorial problem
can be solved efficiently in a polynomial time \cite{Narasimhan2013Structural}.
We briefly discuss their efficient algorithm in this section.

The combinatorial optimization problem associated
with finding the most violated constraint can be written as,
\begin{align}
    \label{EQ:optw1}
        \bar{\bY} =
             \argmax_{\bY \in \sYmn }
    \;
    &Q_{\bw} (\bY),
\end{align}
where
\begin{align}
    \label{EQ:optw2}
        Q_{\bw} (\bY) =
    &\deltapauc(\bYast, \bY) - \\ \notag
     &\frac{1}{mn(\beta-\alpha)}
         \sum_{i,j}  \pi_{ij} \bw^\T (\bh_{:i}^{+} - \bh_{:k_{j}}^{-} ).
\end{align}
The trick to speed up \eqref{EQ:optw1} is to
note that any ordering of the instances that is
consistent with $\bY$ yields the same objective value,
$Q_{\bw} (\bY)$ in \eqref{EQ:optw2}.
In addition, one can break down \eqref{EQ:optw1}
into smaller maximization problems
by restricting the search
space from $\sYmn$ to the set $\sY_{m,\jbeta}^{\bw}$ where
\begin{align}
    \notag
        \sY_{m,\jbeta}^{\bw} =  \bigl\{ \bY \in \sYmn
           | \; \forall i, j_1 < j_2 : \pi_{i,(j_1)_\bw} \geq \pi_{i,(j_2)_\bw} \bigr\}.
\end{align}
Here $\sY_{m,\jbeta}^{\bw}$ represents the set of all matrices $\bY$
in which the ordering of the scores of two negative instances,
$\bw^\T \bh_{:j_1}^{-}$ and $\bw^\T \bh_{:j_2}^{-}$, is consistent.
The new optimization problem is now easier to solve
as the set of negative instances over which the loss term in
\eqref{EQ:optw2} is computed is the same for all
orderings in the search space.
Interested reader may refer to \cite{Narasimhan2013SVM}.
We summarize the algorithm of our \algonametight in Algorithm~\ref{ALG:main}.

A theoretical analysis of the convergence property for Algorithm~\ref{ALG:main}
is as follows.
\begin{prop}
  At each iteration of Algorithm \ref{ALG:main},
  the objective value decreases.
\end{prop}

\begin{prop}
  The decrease of objective value between iterations $ t - 1 $ and $ t $ is not less than
\[
 \Bigl[  \phizeta ( \bh_{t:}, \bYast ) - \phizeta
        ( \bh_{t:}, \bY_{[t]}^\star)
      \Bigr]^2.
\]
    Here,
\begin{equation*}
\bY_{[t]}^\star  =  \argmax_{  \bY  }
  \Bigl\{
  \deltapauc (\bY, \bYast)  +   \bw_{[t]}  ^ \T  \phizeta (\bH, \bY)
\Bigr\},
\end{equation*}
\end{prop}
See supplementary material for the proofs.

\SetKwInput{KwInit}{Initialize}

\begin{algorithm}[t]
\caption{The training algorithm for \algonametight.
}
\begin{algorithmic}
\small{
   \KwIn{
    \\1)    A set of training examples $\{\bx_l,y_l\}$, $l=1, \cdots, m+n$;
     $
     \;
     $
     \\ 2)    The maximum number of weak learners, $\tmax$;
     stopping precision constant $ \varepsilon$;
     $
     \;
     $
     \\ 3)    The regularization parameter, $\nu$;
     $
     \;
     $
     \\ 4)    The learning objective based on the partial AUC, $\alpha$ and $\beta$;
   }

\KwOut{
  The scoring function$^{\dag}$, $f(\bx) = \sum_{t=1}^{\tmax} w_t \hbar_t(\bx)$,
  that optimizes the \pAUC score in the \FPR range $[\alpha, \beta]$;
}

\KwInit {
\\1) $t = 0$;
\\2) Initilaize sample weights: $u_l = \frac{0.5}{m}$ if $y_l = +1$, else $u_l = \frac{0.5}{n}$;
\\3) Extract low level features and store them in the cache memory
for fast data access;
}

\While{ $t < \tmax$ and \eqref{EQ:Stop} is not met}
{
  \cone\ Train a new weak learner using \eqref{EQ:weak2}. The weak learner corresponds to
     the weak classifier with the minimal weighted error (maximal edge);
  \\ \ctwo\ Add the best weak learner into the current set;
  \\ \cthree\ Solve the structured SVM problem using the cutting plane
          algorithm (Algorithm~\ref{ALG:cp});
  \\ \cfour\ Update sample weights, $\bu$, using \eqref{EQ:updatew};
  \\ \cfive\ $t \leftarrow t + 1;$
}

$\dag$ For a node in a cascade classifier, we introduce the threshold,
$b$, and adjust $b$ using the validation set
such that $\sign \bigl( f(\bx) - b \bigr)$ achieves the node learning objective\;

} %
\end{algorithmic}
\label{ALG:main}
\end{algorithm}

\SetKwInput{KwInit}{Initialize}
\SetKwRepeat{Repeat}{Repeat}{Until}

\begin{algorithm}[t]
\caption{The cutting-plane algorithm
}
\begin{algorithmic}
\small{
   \KwIn{
    \\1)      A set of weak learners' outputs $\bH = (\bHp, \bHm)$;
     $
     \;
     $
     \\ 2)    The learning objective based on the partial AUC, $\alpha$ and $\beta$;
     $
     \;
     $
     \\ 3)    The regularization parameter, $\nu$;
     $
     \;
     $
     \\ 4)    The cutting-plane termination threshold, $\epsilon$;
   }

\KwOut{
  The weak learners' coefficients $\bw$, the working set $\sC$ and 
     the dual variables $\blambda$, $\rho$;
}

\KwInit {
$\sC = \varnothing;$
}

$Q_{\bw} (\bY) = \deltapauc(\bYast, \bY) - \frac{1}{mn(\beta-\alpha)}
         \sum_{i,j}  \yij \bw^\T (\bh_{:i}^{+} - \bh_{:k_{j}}^{-} );$

\Repeat{ $ Q_{\bw} (\bYbar) \leq \xi + \epsilon $
}{ 
\ccone\ Solve the dual problem using linear SVM,
\begin{flalign}
    \notag
    \min_{ \bw , \xi } \;
    \frac{1}{2} \| \bw  \|_{2}^{2} + \nu \, \xi  \quad
    \st \;
    Q_{\bw} (\bY) \leq \xi, \forall \bY \in \sC;
\end{flalign}
\cctwo\ Compute the most violated constraint,
\begin{align}
    \notag
    \bYbar = \argmax_{ \bY \in \sY_{m,\jbeta} }  Q_{\bw} (\bY);
\end{align}
\ccthree\ $\sC \leftarrow \sC \cup  \{ \bYbar \};$
}
   
} %
\end{algorithmic}
\label{ALG:cp}
\end{algorithm}

\begin{table*}[t]
\centering
\caption{Computational complexity of our approach and AdaBoost.
$m,n$ are the number of positive and negative training samples,
$F$ is the number of features, $K$ is the number of nodes in the
decision tree, $\tmax$ is the maximum number of weak learners learned
and $r$ is the maximum number of cutting-plane iterations
}
{
\begin{tabular}{l|c|c}
 \hline
   $ $ & \algonametight & AdaBoost \\
 \hline
 \hline
   Training a weak learner (Step \cone\ in Algorithm~\ref{ALG:main}) &
     $\bigO \bigl( (m+n)FK \bigr)$  &
     $\bigO \bigl( (m+n)FK \bigr)$  \\
   Solve $\bw$ (Algorithm~\ref{ALG:cp})  &
     $\bigO \Bigl( r (m+n) \bigl(\log (m+n) + \tmax \bigr)  \Bigr)$ &
     $\bigO (m+n) $ \\
   \hline
   \textbf{Total}  &
     $\bigO \Bigl( \tmax \bigl[ (m+n) (r\log (m+n) + r\tmax + FK ) \bigr] \Bigr)$ &
     $\bigO \Bigl( \tmax \bigl[ (m+n)FK \bigr] \Bigr)$ \\
  \hline
 \end{tabular}
}
\label{tab:complexity}
\end{table*}

\paragraph{Computational complexity}
Each iteration in Algorithm~\ref{ALG:main} consists of $5$ steps.
Step \cone\ learns the weak classifier with the minimal weighted error
and add this weak learner to the ensemble set.
In this step, we train a weak classifier using decision trees.
We train the decision tree using the fast implementation of \cite{Appel2013Quickly},
in which feature values are quantized into $256$ bins.
This procedure costs $\bigO( (m+n)FK  )$ at each iteration.
where $m+n$ is the total number of samples, $F$ is the number of features and
$K$ is the number of nodes in the tree.

We next analyze the time complexity of step \cthree\ which calls
Algorithm~\ref{ALG:cp}.
Algorithm~\ref{ALG:cp} solves the structural SVM problem using
the efficient cutting-plane algorithm.
Step \ccone\ in Algorithm~\ref{ALG:cp} costs $\bigO \bigl(\tmax (m+n) \bigr) $ time since
the linear kernel scales linearly with the number
of training samples \cite{Joachims2006Training}.
Here $\tmax$ is the maximum number of features (weak classifiers).
Using the efficient algorithm of \cite{Narasimhan2013SVM},
step \cctwo\ costs $\bigO \bigl( n \log n + (m+\nbeta) \log (m+\nbeta) \bigr)$
$\leq \bigO \bigl( (m+n) \log (m+n) \bigr)$ time where
$\nbeta = n\beta$ and $\beta \leq 1$.
As shown in \cite{Joachims2009Cutting}, the number of iterations
of Algorithm~\ref{ALG:cp} is upper bounded by the value which is independent
of the number of training samples.
Here, we assume that the number of cutting-plane iterations
required is bounded by $r$.
In total, the time complexity of Algorithm~\ref{ALG:cp} (Step \cthree\
in Algorithm~\ref{ALG:main}) is
$\bigO \Bigl( r \bigl(\log (m+n) + \tmax \bigr) (m+n)  \Bigr)$.
Step \cfour\ updates the sample variables which can be executed
in linear time.
In summary, the total time complexity for training
$\tmax$ boosting iterations using our approach is
$\bigO \Bigl( \tmax \bigl[ (m+n) (r\log (m+n) + r\tmax + FK ) \bigr] \Bigr)$.
From this analysis, most of the training time is spent on training weak
learners when $FK \gg \log(m+n)$.
We summarizes the computational complexity of our approach in
Table~\ref{tab:complexity}.
Table~\ref{tab:complexity} also compares the computational complexity
of our approach with AdaBoost.
We discuss the difference between our approach
and AdaBoost in the next section.

\paragraph{Discussion}
Our final ensemble classifier has a similar form as the
AdaBoost-based object detector of \cite{Viola2004Robust}.
Based on Algorithm~\ref{ALG:main},
step \cone\ and \ctwo\ of our algorithm are identical
to the first two steps of AdaBoost adopted in \cite{Viola2004Robust}.
Similar to AdaBoost, $u_\iprime$ in step \cone\ plays the role of sample weights
associated to each training sample.
The major difference between AdaBoost and our approach is in step \cthree\ and \cfour\,
where the weak learner's coefficient is computed and the sample weights
are updated.
In AdaBoost, the weak learner's coefficient is calculated as
$w_t = \frac{1}{2} \log \frac{1-\epsilon_t}{\epsilon_t}$
where $\epsilon_t = \sum_{l} u_l I \bigl( y_l \neq \hbar_t(\bx_{l}) \bigr)$
and $I$ is the indicator function.
The sample weights are updated with
\[
  u_l = \frac{ u_l \exp (-w_t y_l \hbar_t(\bx_{l}))}
{\sum_{l} u_l \exp (-w_t y_l \hbar_t(\bx_{l})}.
\]
We point this out here since a minimal modification is required in order
to transform the existing implementation of AdaBoost to \algonametight,
due to the high similarity.

We point out here the major difference between the ensemble classifier
proposed in this paper and our earlier work \cite{Paul2013Efficient}.
In this work, we redefine the joint feature map \eqref{EQ:featmap}
over subsets of negative instances.
The new feature map leads to a tighter hinge relaxation on the partial AUC loss.
In other words, the joint feature map defined in \cite{Paul2013Efficient}
is computed over all the negative instances instead of subsets of negative
instances ranked in positions ${j_\alpha + 1, \cdots, j_\beta}$
(corresponding to the \FPR range $[\alpha, \beta]$ one is interested in).
As a result, the new formulation is not only faster to train
but also perform slightly better on the pAUC measure than \cite{Paul2013Efficient}.

\subsection{Region proposals generation}
The evaluation of our pedestrian detector outlined in the previous subsections can be
computed efficiently with the use of integral channel features \cite{Dollar2009Integral}.
However the detector is still not efficient enough to be used in a sliding-window based framework.
In order to improve the evaluation time, we adopt a cascaded approach in which
classifiers are arranged based on their complexity \cite{Viola2004Robust}.
In this paper, we adopt a two-stage approach, in which we place the
fast to extract features in the first stage and our proposed features
with \algonametight in the second stage.
In other words, our proposed detector is evaluated only on test samples
which pass through the first stage.

Recently the binarized normed gradients (BING) feature with a linear SVM classifier
has been shown to speed up the classical sliding window object detection paradigm by discarding
a large set of background patches \cite{Cheng2014Bing}.
On Pascal VOC2007, it achieves a comparable object detection rate to recently proposed
Objectness \cite{Alexe2012Measuring} and Selective Search \cite{Uijlings2013Selective},
while being three orders of magnitudes faster than these approaches.
The detector of \cite{Cheng2014Bing} is adopted in the first stage of our two-stage detector
to filter out a large number of background patches.
The underlying idea is to reduce the number of samples that our proposed detector needs to process.

\paragraph{Implementation}
The original BING detector of \cite{Cheng2014Bing} was trained for generic object detection.
We make the following modifications to the original BING detector to improve its performance for pedestrian detection.
\begin{enumerate}
\item Instead of resizing the training data to a resolution of $8 \times 8$ pixels,
we resize the resolution of pedestrian samples to $8 \times 16$ pixels.
This template has the same aspect ratio as the one adopted in \cite{Dalal2005HOG}.
Hence the BING features we use in our paper is $128$ bit integer instead
of $64$ bit integer used in the original paper.
The data type \UINT128 is adopted to store BING features.
In addition, we replace the Pascal VOC2007 training data with
the Caltech training data to train the BING detector.

\item The original paper quantizes the test image to a resolution $\{(w,h)\}$
where $w,h \in \{10,$ $20,$ $40,$ $80,$ $160,$ $320\}$.
In this paper, we apply a multi-scale detection with a fixed aspect ratio.
We scan the test image at $8$ scales per octave (corresponding to a scale stride of $1.09$).
\end{enumerate}

\section{Experiments}
\label{sec:exp}

\subsection{Spatially pooled features}
We compare the performance of the proposed feature
with and without spatial pooling.
Our \SCOV consists of $9$ low-level image statistics.
We exclude the mean and variance of two image statistics (pixel locations at x and y co-ordinates)
since they do not capture discriminative information.
We also exclude the correlation coefficient between pixel locations at x and y co-ordinates.
Hence there is a total of $136$ channels ($7$ low-level image statistics +
$3 \cdot 7$ variances + $3 \cdot 35$ correlation coefficients + $3$ LUV color
channels)\footnote{Note here that we extract
covariance features at $3$ different scales.}.
Experiments are carried out using AdaBoost with the shrinkage parameter of
$0.1$ \cite{Hastie2009Elements} and
level-$3$ decision trees as weak classifiers.
We apply shrinkage to AdaBoost as it has been shown to improve the final
classification accuracy \cite{Friedman2000Additive}.
We use the depth-3 decision tree as it offers better generalization performance
as shown in \cite{Paul2014Strengthening}.
We train three bootstrapping iterations and the final
model consists of $2048$ weak classifiers with soft cascade.
We heuristically set the soft cascade's rejection threshold to be $-10$ at every node.
Log-average miss rates of detectors trained
using covariance descriptors and LBP
(without and with spatial pooling)
are shown in Table~\ref{tab:pooling}.
We observe that it is beneficial to apply spatial pooling as
it increases the robustness of the features against
small deformations and translations.
We observe a reduction in miss rate by more than one percent on the INRIA test set.
Since we did not combine \SLBP with HOG as in \cite{Wang2009HOG},
\SLBP performs slightly worse than \SCOV.

\begin{table}[bt]
  \caption{
  Log-average miss rate of our features with and without
  applying spatial pooling.
  We observe that spatial pooling improves the translation invariance
  of our features
  }
  \centering
  \scalebox{1}
  {
  \begin{tabular}{l|cc|cc}
  \hline
    \multirow{2}{*}{Data set} & \multicolumn{2}{c|}{\SCOV} &  \multicolumn{2}{c}{\SLBP} \\
  \cline{2-5}
     & without & with pooling & without & with pooling \\
  \hline
  \hline
    INRIA \cite{Dalal2005HOG}  & $14.2\%$ & $\mathbf{12.8\%}$ & $23.2\%$ & $\mathbf{21.8\%}$ \\
    ETH \cite{Ess2008Mobile}    & $42.7\%$ & $\mathbf{42.0\%}$ & $47.2\%$ & $\mathbf{47.1\%}$ \\
    TUD-Br. \cite{Wojek2009Multi} & $48.6\%$  & $\mathbf{47.8\%}$ & $54.9\%$ & $\mathbf{54.4\%}$\\
  \hline
  \end{tabular}
  }
  \label{tab:pooling}
\end{table}

\begin{figure}[t]
    \centering
        \includegraphics[width=0.4\textwidth,clip]{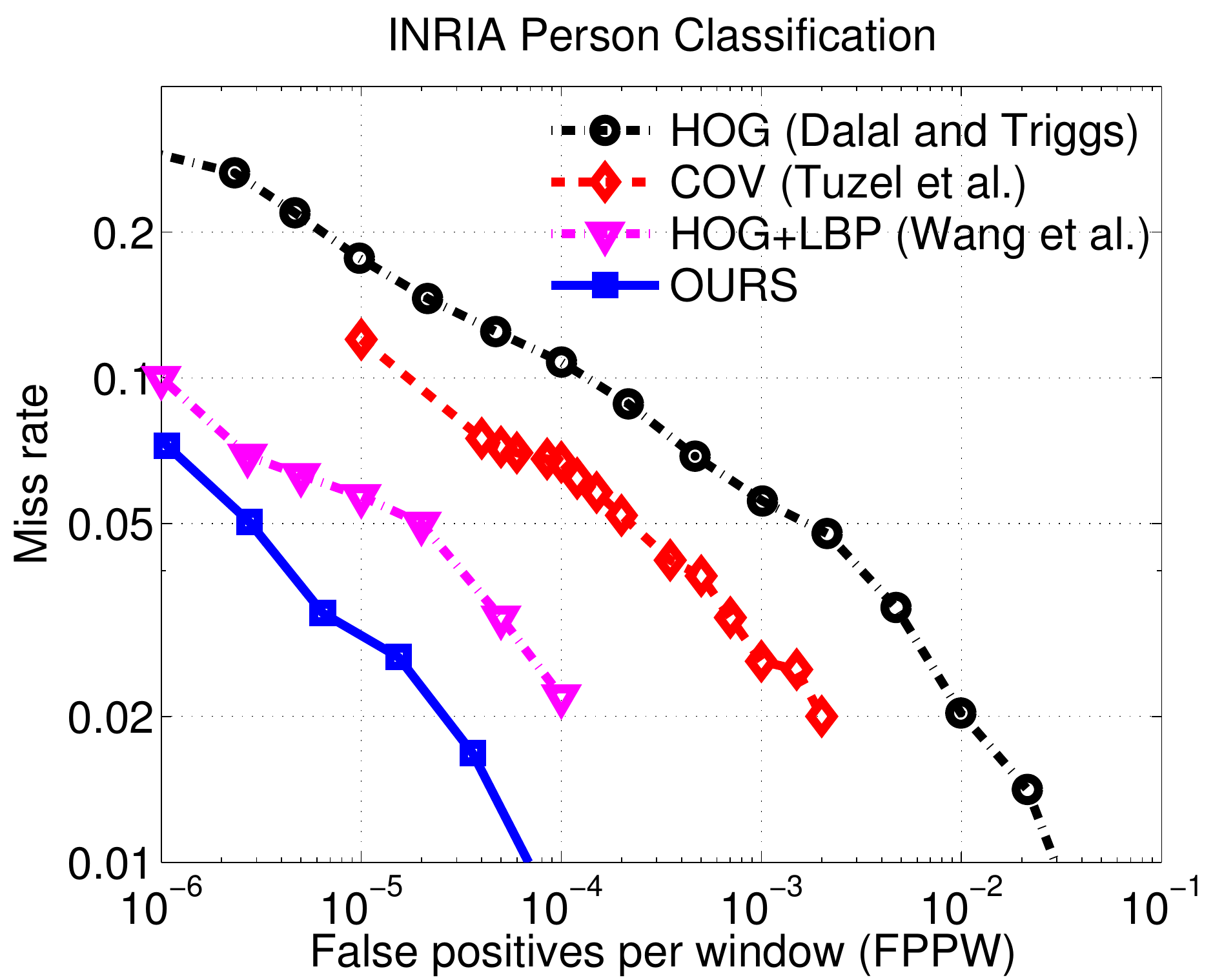}
    \caption{
    ROC curves of our \SCOV features and
    the conventional covariance detector \cite{Tuzel2008Pedestrian}
    on INRIA test images.
    }
    \label{fig:cov_fppw}
\end{figure}

\paragraph{Compared with other pedestrian detectors}
In this experiment, we compare the performance of our proposed \SCOV
with the original covariance descriptor proposed in \cite{Tuzel2008Pedestrian}.
\cite{Tuzel2008Pedestrian} calculates the covariance distance in the Riemannian manifold.
As eigen-decomposition is performed, the approach of \cite{Tuzel2008Pedestrian} is computationally expensive.
We speed up the weak learner training by proposing our modified covariance features and
train the weak learner using the decision tree.
The new weak learner is not only simpler than \cite{Tuzel2008Pedestrian} but also highly effective.
We compare our previously trained detector with
the original covariance descriptor \cite{Tuzel2008Pedestrian} in Fig.~\ref{fig:cov_fppw}.
We plot HOG \cite{Dalal2005HOG} and HOG+LBP \cite{Wang2009HOG} as the baseline.
Similar to the result reported in \cite{Benenson2013Seeking}, where the authors show that
\HOGBOOST reduces the average miss-rate over \HOGSVM by more than $30\%$,
we observe that applying our \SCOV features as the channel features
significantly improves the detection performance over the original covariance detector
(a reduction of more than $5\%$ miss rate at $10^{-4}$ false positives per window).

Next we compare the proposed \SCOV with \ACF features (M+O+LUV) \cite{Dollar2014Fast}.
Since \ACF uses fewer channels than \SCOV,
for a fair comparison, we increase \nACF's discriminative power by combining
\ACF features with LBP\footnote{In our implementation, we use an extension of
LBP, known as the uniform LBP, which can better filter out noises \cite{Wang2009HOG}.
Each LBP bin corresponds to each channel.} (M+O+LUV+LBP).
The results are reported in Table~\ref{tab:featcomb}.
We observe that \SCOV yields competitive results to M+O+LUV+LBP.
From the table, \SCOV performs better on the INRIA test set, worse on the ETH test set and on par with
M+O+LUV+LBP on the TUD-Brussels test set.
We observe that the best performance is achieved by combining \SCOV and \SLBP with
M+O+LUV.

\begin{table}[tb]
  \caption{
  Log-average miss rates of various feature combinations
  }
  \centering
  \scalebox{1}
  {
  \begin{tabular}{l|ccc}
  \hline
     & INRIA & ETH & TUD-Br. \\
  \hline
  \hline
    M+O+LUV+LBP        & $14.5\%$ & $39.9\%$ & $47.0\%$ \\
    \nSCOV+LUV         &$12.8\%$ & $42.0\%$ & $47.8\%$ \\
    \nSCOV+M+O+LUV     & $\mathbf{11.2\%}$ & $39.4\%$ & $46.7\%$ \\
    \nSCOV+\nSLBP+M+O+LUV  & $\mathbf{11.2\%}$ & $\mathbf{38.0\%}$ & $\mathbf{42.5\%}$ \\
  \hline
  \end{tabular}
  }
  \label{tab:featcomb}
\end{table}

 \begin{figure*}[t]
        \centering
        \includegraphics[width=0.245\textwidth,clip]{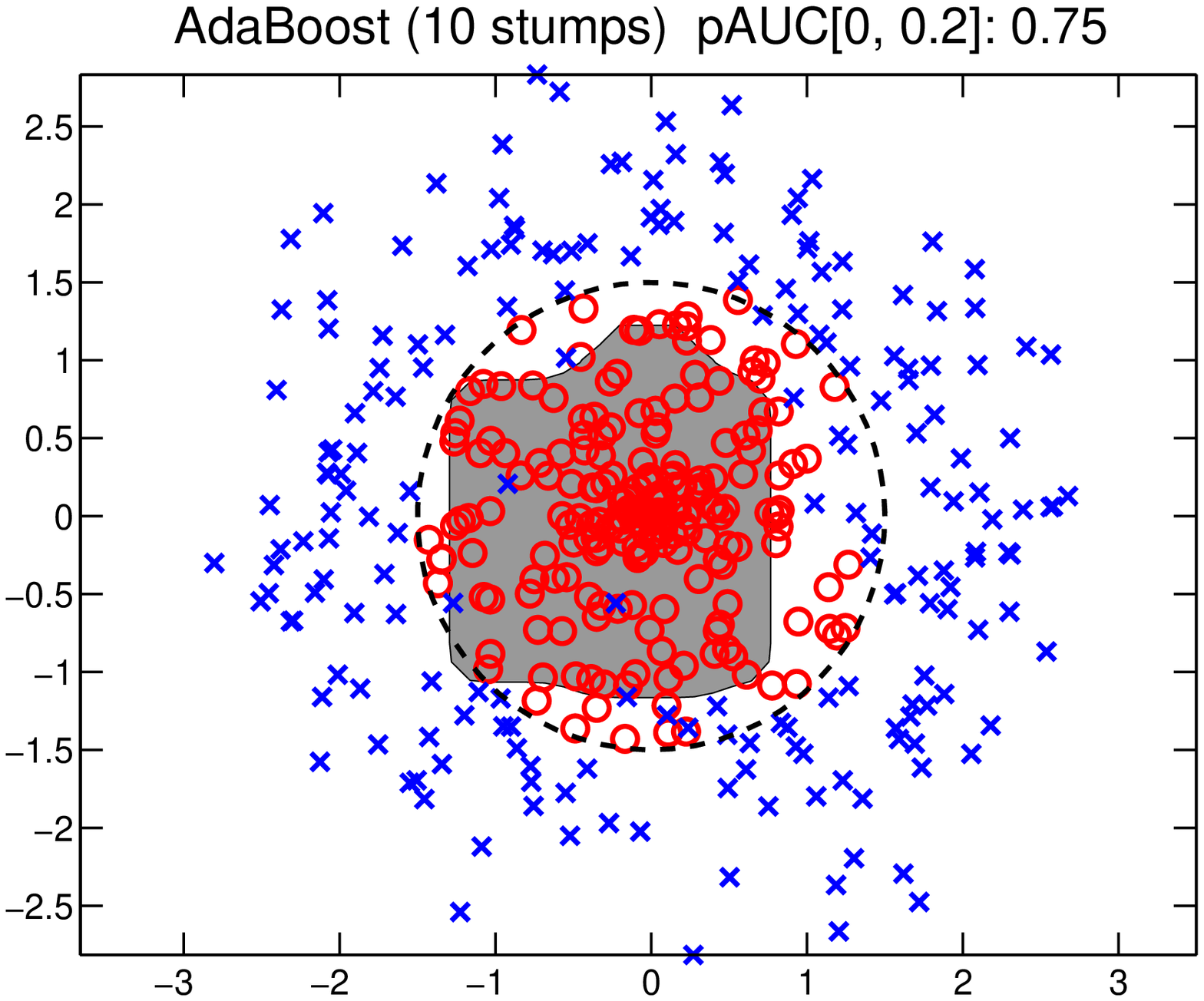}
        \includegraphics[width=0.245\textwidth,clip]{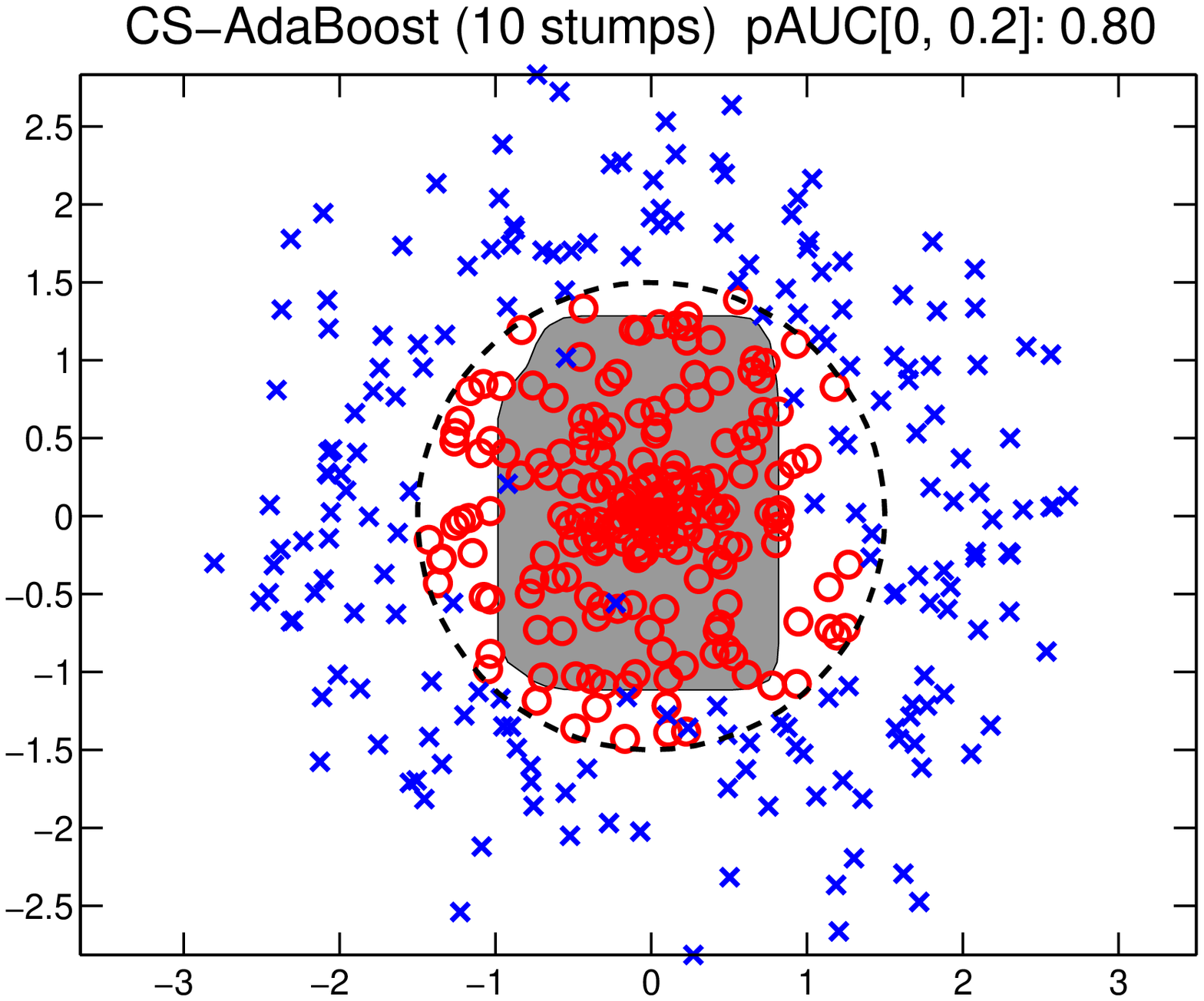}
        \includegraphics[width=0.245\textwidth,clip]{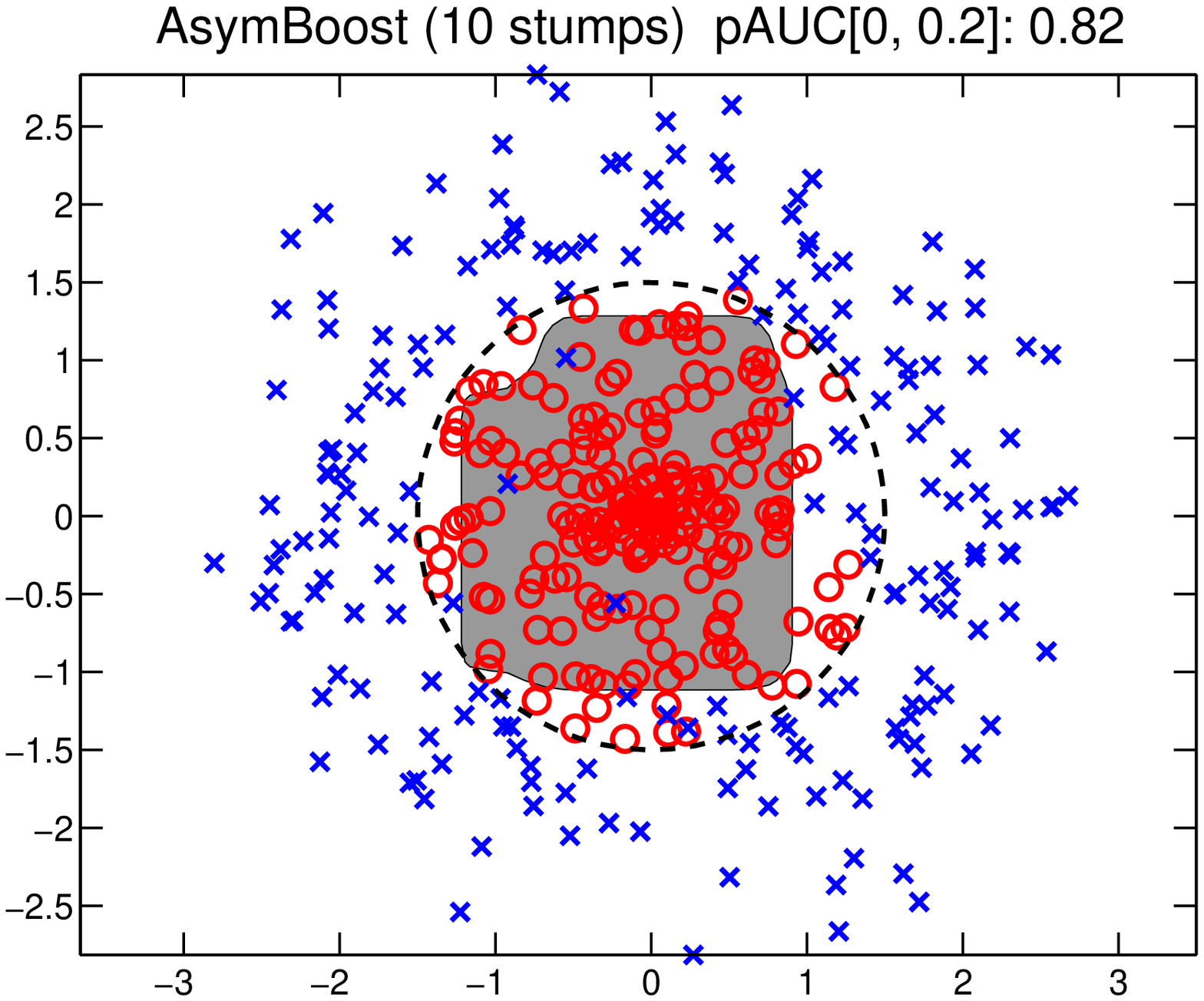}
        \includegraphics[width=0.245\textwidth,clip]{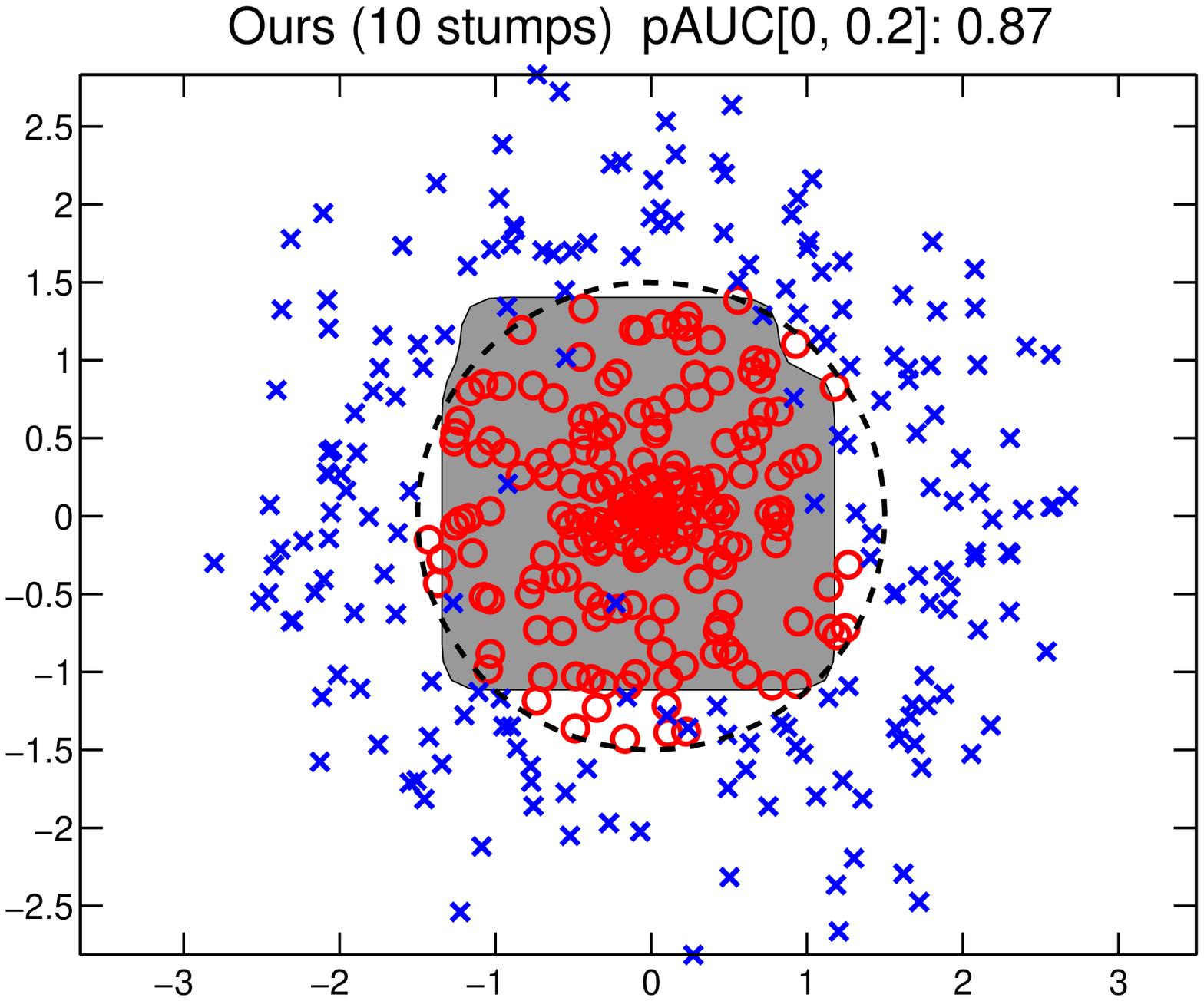}
    \caption{
    Decision boundaries on the toy data set where each strong classifier consists of
    $10$ weak classifiers (horizontal and vertical decision stumps).
    Positive and negative data are represented by $\circ$ and $\times$, respectively.
    The partial AUC score in the \FPR range $[0, 0.2]$ is also displayed.
    Our approach achieves the best pAUC score compared to other asymmetric classifiers.
    }
    \label{fig:toy1}
  \end{figure*}

\subsection{Ensemble classifier}
\paragraph{Synthetic data set}
In this experiment, we illustrate the effectiveness of our ensemble classifier
on a synthetic data set similar to the one evaluated in \cite{Viola2002Fast}.
The radius and angle of the positive data is drawn from a uniform distribution
$[0, 1.5]$ and $[0, 2 \pi ]$, respectively.
The radius of the negative data is drawn from a normal distribution with
mean of $2$ and the standard deviation of $0.4$.
The angle of the negative data is drawn from a uniform distribution similar to
the positive data.
We generate $400$ positive data and $400$ negative data for
training and validation purposes
($200$ for training and $200$ for validating the asymmetric parameter).
For testing, we evaluate the learned classifier with
$2000$ positive and negative data.
We compare \algonametight against the baseline AdaBoost,
Cost-Sensitive AdaBoost (CS-AdaBoost) \cite{Masnadi2011Cost} and
Asymmetric AdaBoost (AsymBoost) \cite{Viola2002Fast}.
For CS-AdaBoost, we set the cost for misclassifying positive
and negative data as follows.
We assign the asymmetric factor $k = C_1 / C_2$ and restrict
$0.5(C_1 + C_2) = 1$.
We then choose the best $k$ which returns the highest partial AUC
from $\{0.5$, $0.6$, $\cdots$, $2.4$, $2.5\}$.
For AsymBoost, we choose the best asymmetric factor $k$ which
returns the highest partial AUC from $\{2^{-1}$, $2^{-0.8}$, $\cdots$,
$2^{1.8}$, $2^{2}\}$.
For our approach, the regularization parameter is chosen from
$\{10^{-5}$, $10^{-4.8}$, $\cdots$, $10^{-3.2}$, $10^{-3}\}$.
We use vertical and horizontal decision stumps as the weak classifier.
For each algorithm, we train a strong classifier consisting of $10$ weak classifiers.
We evaluate the partial AUC of each algorithm at $[0, 0.2]$ FPRs.

Fig.~\ref{fig:toy1} illustrates the boundary
decision\footnote{We set the threshold such that
the false positive rate is $0.2$.}
and the pAUC score.
Our approach outperforms all other asymmetric classifiers.
We observe that \algonametight places more emphasis on
positive samples than negative samples to ensure
the highest detection rate at the left-most part of the
ROC curve (FPR $< 0.2$).
Even though we choose the asymmetric parameter, $k$, from a large range of
values, both CS-AdaBoost and AsymBoost perform slightly worse than our
approach.
AdaBoost performs worst on this toy data set
since it optimizes the overall classification accuracy.
\begin{table}[bt]
  \caption{The pAUC score on Protein-protein interaction data set.
  The \textbf{higher the pAUC score}, the \textbf{better the classifier}.
  Results marked by $\dag$ were reported in \cite{Narasimhan2013Structural}.
  The best classifier is shown in boldface
  }
  \centering
  {
  \begin{tabular}{l|c}
  \hline
    & pAUC($0$, $0.1$)  \\
  \hline
  \hline
   Ours (\algonametight) & $\mathbf{56.76\%}$ \\
   \algoname $[0,0.1]$ \cite{Paul2013Efficient} & $56.05\%$ \\
   SVM$_{\rm pAUC}^{\rm tight}$ $[0,0.1]$ \cite{Narasimhan2013SVM} & $52.95\%$ \\
   SVM$_{\rm pAUC}^{\rm struct}$ $[0,0.1]$ \cite{Narasimhan2013Structural} & $51.96\%$ \\
   pAUCBoost $[0,0.1]^{\dag}$ \cite{Komori2010Boosting} & $48.65\%$ \\
   Asym SVM $[0,0.1]^{\dag}$ \cite{Wu2008Asymmetric} & $44.51\%$ \\
   SVM$_{\rm AUC}^{\dag}$ \cite{Joachims2009Cutting} & $39.72\%$ \\
  \hline
  \end{tabular}
  }
  \label{tab:ppi}
\end{table}

\paragraph{Protein-protein interaction prediction}
In this experiment, we compare our approach with
existing algorithms which optimize \pAUC in bioinformatics.
The problem we consider here is a protein-protein interaction
prediction \cite{Qi2006Evaluation}, in which the task is to predict
whether a pair of proteins interact or not.
We used the data set labelled `Physical Interaction Task
in Detailed feature type', which is publicly available
on the internet\footnote{\url{http://www.cs.cmu.edu/~qyj/papers_sulp/proteins05_pages/feature-download.html}}.
The data set contains $2865$ protein pairs known to be interacting (positive)
and a random set of $237,384$ protein pairs labelled as non-interacting (negative).
We use a subset of $85$ features as in \cite{Narasimhan2013Structural}.
We randomly split the data into two groups: $10\%$ for training/validation
and $90\%$ for evaluation.
We choose the best regularization parameter form $\{1$, $1/2$, $1/5\}$ by $5$-fold cross validation.
We repeat our experiments $10$ times using the same regularization parameter.
We train a linear classifier as our weak learner using
\textsc{Liblinear}  \cite{Fan2008Liblinear}.
We set the maximum number of boosting iterations to $100$
and report the pAUC score of our approach in Table~\ref{tab:ppi}.
Baselines include \algoname, SVM$_{\rm pAUC }$,
SVM$_{\rm AUC }$, pAUCBoost and Asymmetric SVM.
Our approach outperforms all existing algorithms
which optimize either \AUC or \pAUC.
We attribute our improvement over
SVM$_{\rm pAUC}^{\rm tight}$ $[0,0.1]$ \cite{Narasimhan2013SVM}, as a result of introducing
a non-linearity into the original problem.
This phenomenon has also been observed in face detection
as reported in \cite{Wu2008Fast}.

\begin{table}[tb]
  \centering
  \caption{Average pAUC scores in the \FPR range $[0, 0.1]$
  and their standard deviations
  on vision data sets at various boosting iterations.
  Experiments are repeated $20$ times.
  The best average performance is shown in boldface
  }
  {
  \begin{tabular}{c|c|ccc}
  \hline
 &  $\#$ iters & USPS  & SCENE  & FACE  \\
\hline
\hline
  Ours & $10$ & $\mathbf{0.88}$ ($\mathbf{0.01}$)  & $\mathbf{0.72}$ ($\mathbf{0.03}$)  & $\mathbf{0.72}$ ($\mathbf{0.02}$) \\
       & $20$ & $\mathbf{0.92}$ ($\mathbf{0.01}$)  & $\mathbf{0.78}$ ($\mathbf{0.03}$)  & $\mathbf{0.81}$ ($\mathbf{0.01}$) \\
   & $100$ & $\mathbf{0.97}$ ($\mathbf{0.00}$)  & $\mathbf{0.87}$ ($\mathbf{0.02}$)  & $\mathbf{0.91}$ ($\mathbf{0.01}$) \\
\hline
  AdaBoost  & $10$ & $0.87$ ($0.01$)  & $0.70$ ($0.03$)  & $0.71$ ($0.02$) \\
  \cite{Viola2004Robust} & $20$ & $0.91$ ($0.01$)  & $0.77$ ($0.03$)  & $0.79$ ($0.01$) \\
   & $100$ & $0.96$ ($0.01$)  & $0.85$ ($0.02$)  & $0.90$ ($0.01$) \\
\hline
  Ada + LDA & $10$ & $0.87$ ($0.02$)  & $0.70$ ($0.03$)  & $0.71$ ($0.02$) \\
  \cite{Wu2008Fast} & $20$ & $0.91$ ($0.01$)  & $0.77$ ($0.03$)  & $0.80$ ($0.01$) \\
   & $100$ & $0.96$ ($0.01$)  & $0.85$ ($0.02$)  & $0.90$ ($0.01$) \\
\hline
  AsymBoost & $10$ & $0.87$ ($0.02$)  & $0.71$ ($0.03$)  & $\mathbf{0.72}$ ($\mathbf{0.02}$) \\
  \cite{Viola2002Fast} & $20$ & $0.91$ ($0.01$)  & $0.77$ ($0.03$)  & $0.79$ ($0.01$) \\
   & $100$ & $0.96$ ($0.00$)  & $0.85$ ($0.02$)  & $0.90$ ($0.01$) \\
\hline
  \end{tabular}
  }
  \label{tab:expvision}
\end{table}

\paragraph{Comparison to other asymmetric boosting}
Here we compare \algonametight against
existing asymmetric boosting algorithms, namely,
AdaBoost with Fisher LDA post-processing \cite{Wu2008Fast} and
AsymBoost \cite{Viola2002Fast}.
The results of AdaBoost are also presented as the baseline.
For each algorithm, we train a strong classifier consisting of
$100$ weak classifiers (decision trees of depth $2$).
We then calculate the pAUC score by varying the threshold value
in the \FPR range $[0, 0.1]$.
For each algorithm, the experiment is repeated $20$ times and
the average pAUC score is reported.
For AsymBoost, we choose $k$ from  $\{2^{-0.5}$, $2^{-0.4}$, $\cdots$,
$2^{0.5} \}$ by cross-validation.
For our approach, the regularization parameter is chosen from
$\{1$, $0.5$, $0.2$, $0.1\}$ by cross-validation.
We evaluate the performance of all algorithms on $3$
vision data sets: USPS digits, scenes and face data sets.
For USPS, we use raw pixel values and
categorize the data sets into two classes:
even digits and odd digits.
For scenes, we divide the $15$-scene data sets used in
\cite{Lazebnik2006Beyond} into $2$ groups: indoor and outdoor scenes.
We use CENTRIST as our feature descriptors and
build $50$ visual code words using the histogram intersection kernel \cite{Wu2011CENTRIST}.
Each image is represented in a spatial hierarchy manner.
Each image consists of $31$ sub-windows.
In total, there are $1550$ feature dimensions per image.
For faces, we use face data sets from \cite{Viola2004Robust} and
randomly extract $5000$ negative patches from background images.
We apply principle component analysis (PCA) to preserve $95\%$ total variation.
The new data set has a dimension of $93$.
We report the experimental results in Table~\ref{tab:expvision}.
From the table, \algonametight demonstrates the best performance on all
three vision data sets.

\begin{figure*}[t]
        \centering
        \includegraphics[width=0.32\textwidth,clip]{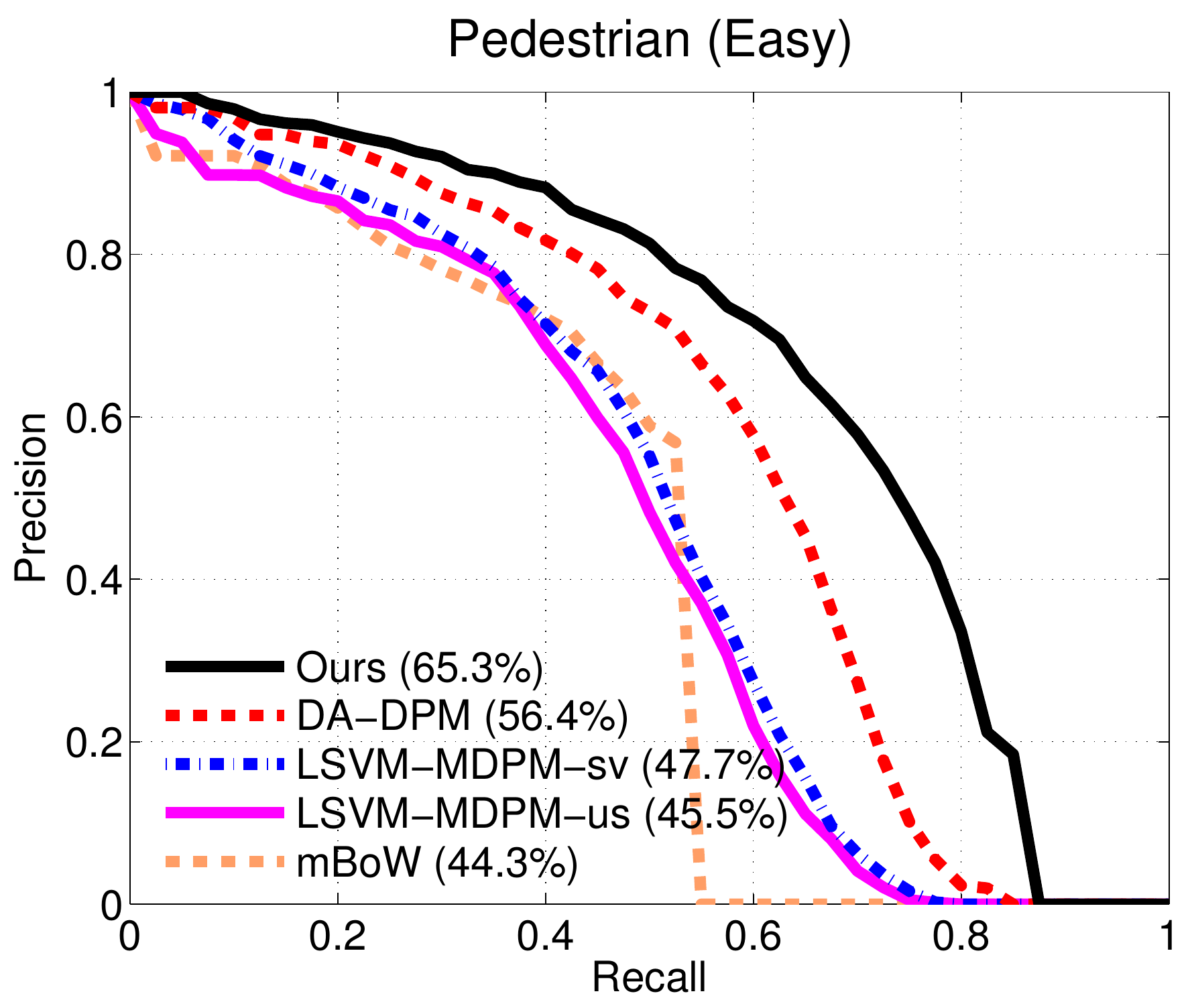}
        \includegraphics[width=0.32\textwidth,clip]{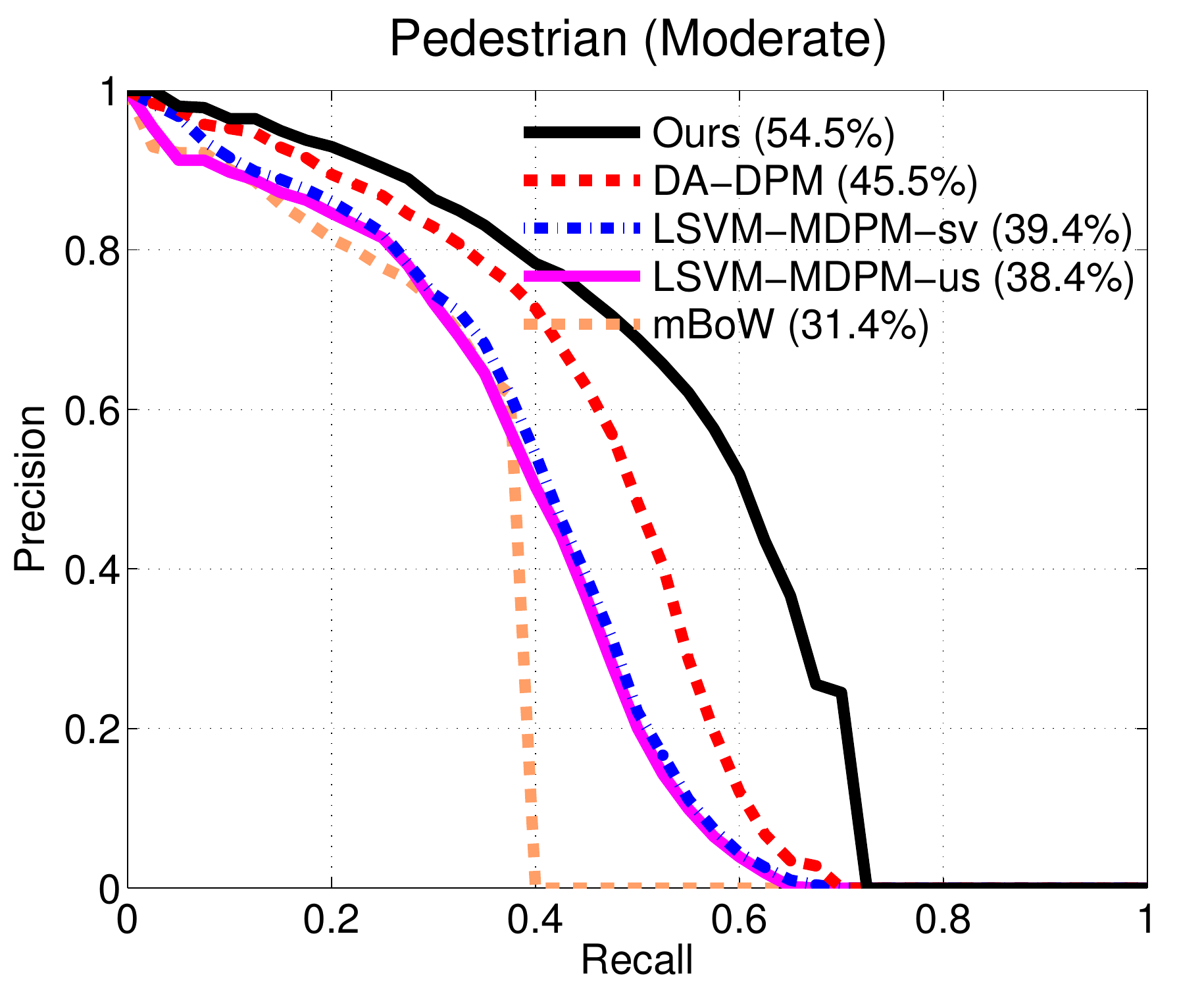}
        \includegraphics[width=0.32\textwidth,clip]{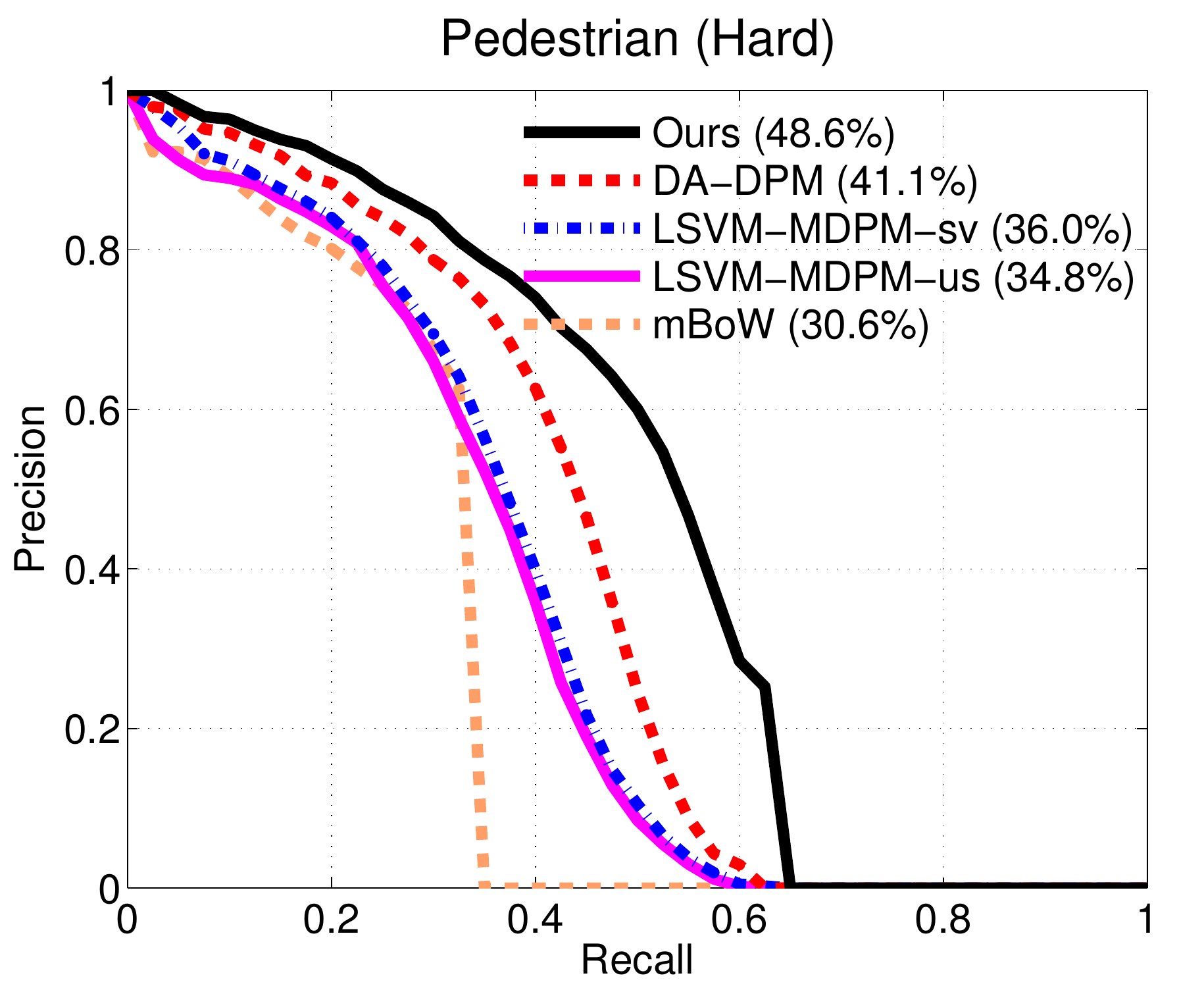}
    \caption{
    Precision-recall curves of our approach and state-of-the-art
    detectors (DA-PDM \cite{Xu2014Hierarchical}, 
    LSVM-MDPM-sv \cite{Urtasun2011Joint},
    LSVM-MDPM-us \cite{Felzenszwalb2010Object} and
    mBoW \cite{Behley2013Laser})
    on the KITTI pedestrian detection test set.
    }
    \label{fig:KITTI}
\end{figure*}

\begin{figure}[t]
        \centering
        \includegraphics[width=0.5\textwidth,clip]{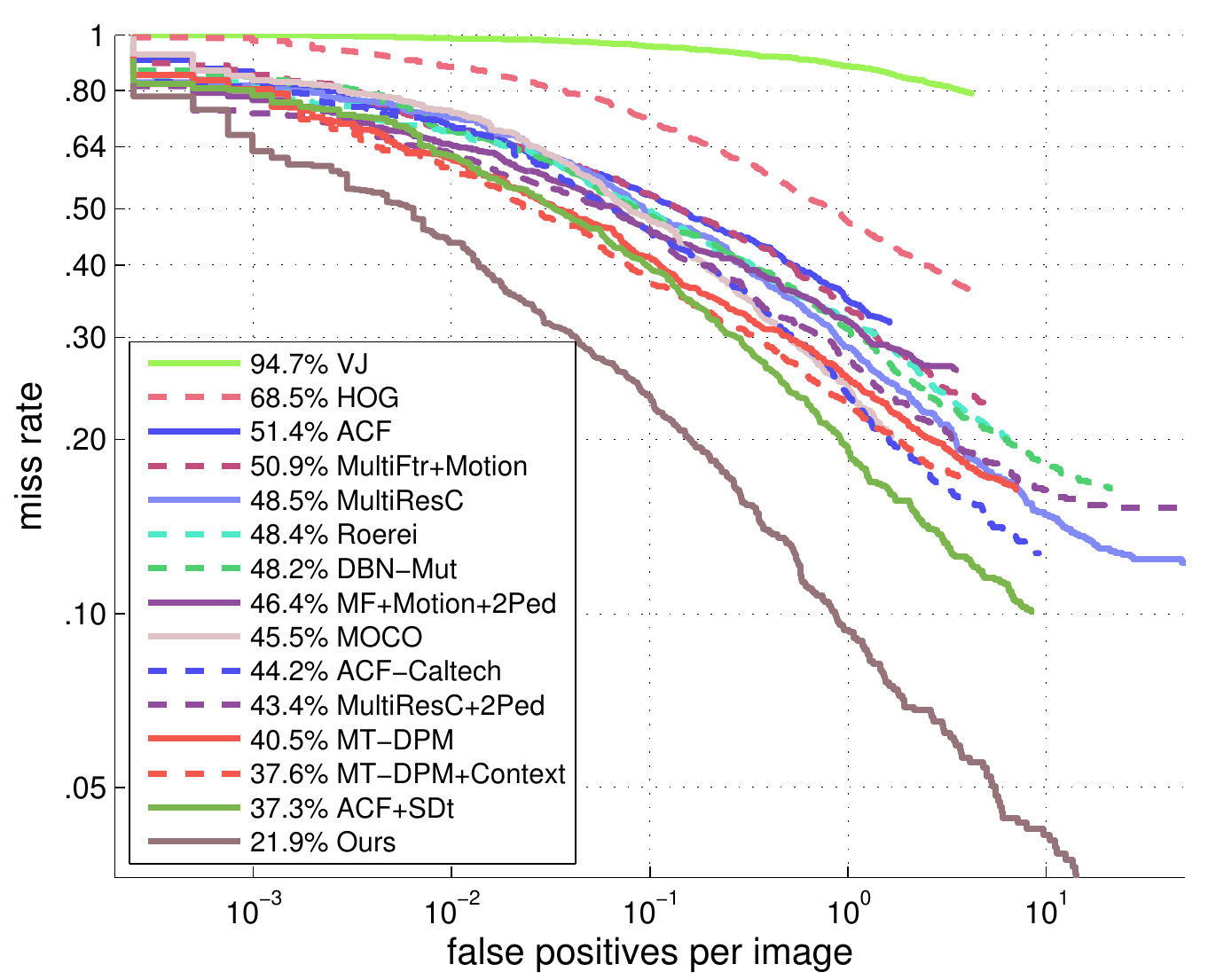}
    \caption{ROC curves of our approach and several state-of-the-art
    detectors (ACF+SDt \cite{Park2013Exploring},
    MT-DPM+Context \cite{Yan2013Robust},
    MT-DPM \cite{Yan2013Robust},
    MultiResC+2Ped \cite{Ouyang2013Single},
    ACF-Caltech \cite{Dollar2014Fast},
    MOCO \cite{Chen2013Detection},
    MF+Motion+2Ped \cite{Ouyang2013Single},
    DBN-Mut \cite{Ouyang2013Modeling},
    Roerei \cite{Benenson2013Seeking},
    MultiResC \cite{Park2010Multiresolution},
    MultiFtr+Motion \cite{Walk2010New},
    ACF \cite{Dollar2014Fast},
    HOG \cite{Dalal2005HOG} and
    VJ \cite{Viola2004Robust})
    on the Caltech pedestrian test set.
}
    \label{fig:strong1}
\end{figure}

\begin{figure}[t]
        \centering
        \includegraphics[width=0.154\textwidth,clip]{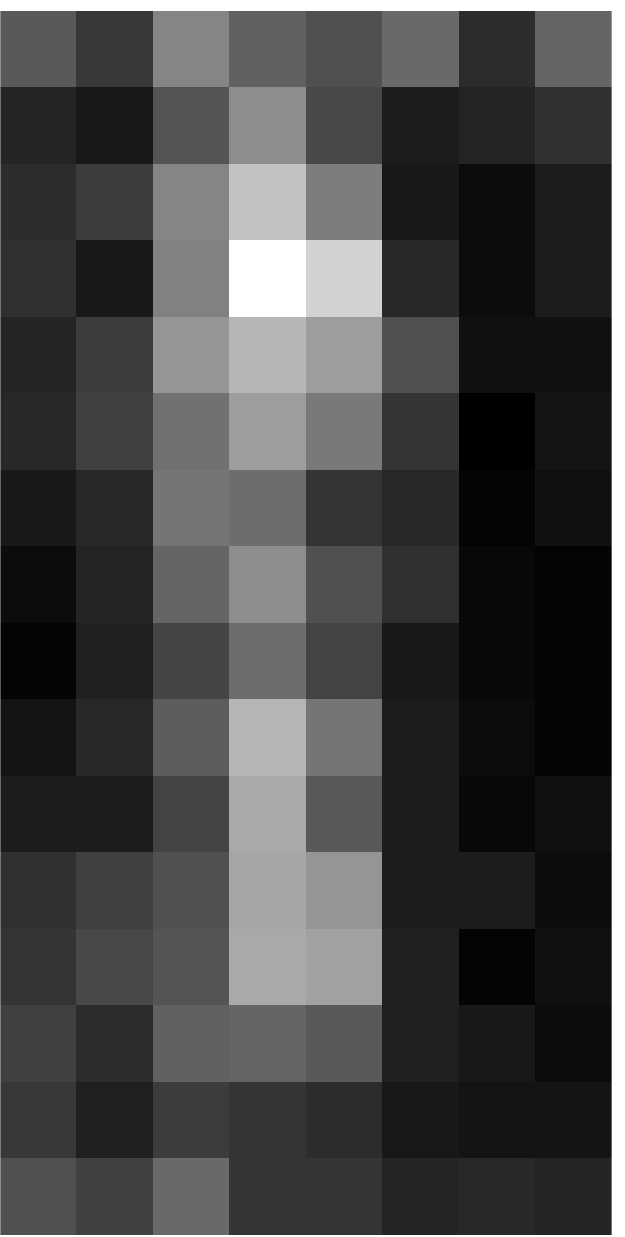}
        \includegraphics[width=0.15\textwidth,clip]{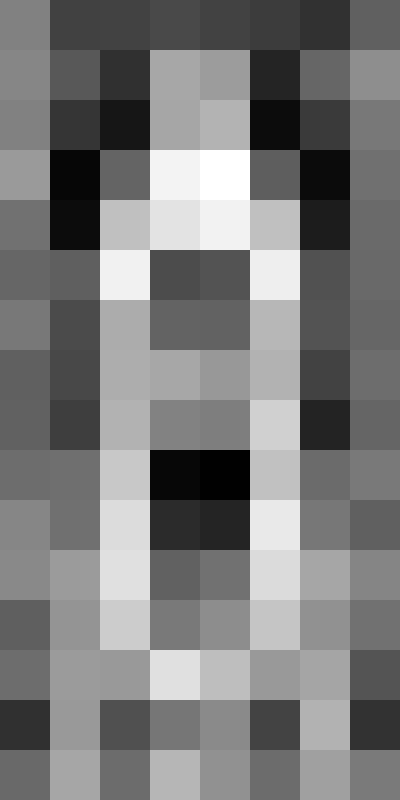}
    \caption{
    {\em Left:} Spatial distribution of features
    selected by \algonametight.
    White pixels indicate that a large number of low-level
    visual features are selected in that area.
    These regions correspond to human head, shoulders and feet.
    {\em Right:} The learned linear SVM model from the BING classifier.
    Each pixel shows the SVM weight.
    Note the similarity between the learned SVM weights
    and SVM weights of HOG (Fig.~6b in \cite{Dalal2005HOG}), \ie,
    large SVM weights are near the head
    and shoulder contour ($\wedge$-shape).
    }
    \label{fig:filter}
\end{figure}

\subsection{Pedestrian detection}
\label{sec:exp_ped}
We evaluate the performance of our approach on the pedestrian detection task.
We train the pedestrian detector on the KITTI vision 
benchmark suite \cite{Geiger2012Kitti} and
Caltech-USA pedestrian data set \cite{Dollar2012Pedestrian}.
The KITTI data set was captured from two high resolution stereo 
camera systems mounted to a vehicle.
The vehicle was driven around a mid-size city.
All images are color and saved in the 
portable network graphics (PNG) format.
The data set consists of $7481$ training images and
$7518$ test images.
To obtain positive training data for the KITTI data set, 
we crop $2111$ fully visible pedestrians
from $7481$ training images.
We expand the positive training data by flipping cropped pedestrian patches
along the vertical axis.
Negative patches are collected from the KITTI training set with
pedestrians, cyclists and `don't care' regions cropped out.
To train the pAUC-based pedestrian detector, we set
the resolution of the pedestrian model to $32 \times 64$ pixels.
We extract visual features based on integral channel features approach.
We use five different types of features: color (LUV), magnitude,
orientation bins \cite{Dollar2009Integral}, the proposed \SCOV
and the proposed \SLBP.
We use decision trees as weak learners and set the depth of decision trees to be three.
The regularization parameter $\nu$ is cross-validated from $\{1, 2^{-1}, \cdots, 2^{-4}\}$
using the KITTI training set (dividing the training set into training and validation splits).
For \FPR range $[\alpha,\beta]$, we set the $\alpha$ to $0$ and
again choose the value of $\beta$ from $\{1, 2^{-1}, \cdots, 2^{-4}\}$ on the cross-validation data set.
The \algonametight detector is repeatedly trained with three bootstrapping iterations and
the total number of negative samples collected is around $33,000$.
The final classifier consists of $2048$ weak classifiers.
To obtain final detection results, greedy non-maxima suppression is applied
with the default parameter as described in the Addendum of \cite{Dollar2009Integral}.
We submit our detection results to the KITTI benchmark website
and report the precision-recall curves of our detector in Fig.~\ref{fig:KITTI}.
{\em The proposed approach outperforms all existing pedestrian detectors reported
so far on the KITTI benchmark website.}

Next, we evaluate our classifier on the Caltech-USA benchmark data set.
The Caltech-USA data set was collected from a video taken 
from a vehicle driving through regular traffic in an urban 
environment (greater Los Angeles metropolitan area).
Images were captured at a resoltuion of $640 \times 480$ pixels
at $30$ frames per second.
Pedestrian are categorized into three scales.
The near scale includes pedestrians over $80$ pixels,
the medium scale includes pedestrians at $30-80$ pixels
and the far scale includes pedestrian under $30$ pixels.
In this paper, we use every $30^{\textrm{th}}$ frame 
(starting with the $30^{\textrm{th}}$ frame) of the Caltech dataset.
The data is split into training and test sets.
For the positive training data,
we use $1631$ cropped pedestrian patches extracted from
$4250$ training images.
We exclude occluded pedestrians from the Caltech training set \cite{Park2013Exploring}.
Pedestrian patches are horizontally mirrored (flipped along the vertical axis)
to expand the positive training data.
Negative patches are collected from the Caltech-USA training set with
pedestrians cropped out.
To train the pAUC-based pedestrian detector, we set
the resolution of the pedestrian model to $32 \times 64$ pixels.
We use six different types of features: color (LUV), magnitude,
orientation bins \cite{Dollar2009Integral},
histogram of flow\footnote{
We use the optical flow implementation of \cite{Liu2009Beyond}
which can be downloaded at \url{http://people.csail.mit.edu/celiu/OpticalFlow/}
} \cite{Dalal2006Human}, \SCOV and \SLBP.
We set the depth of decision trees, the number of bootstrapping iterations and
the number of weak classifiers to be the same as in the previous experiment.
We evaluate our pedestrian detectors on the
conditions that pedestrians are at least $50$ pixels in height
and at least $65\%$ visible.
We use the publicly available evaluation software
of Doll\'{a}r \etal \cite{Dollar2012Pedestrian},
which computes the AUC from $9$ discrete points sampled between $[0.01, 1.0]$ \nFPPI,
to evaluate our experimental results.
Fig.~\ref{fig:strong1} compares the performance of
our approach with other state-of-the-art algorithms.

{\em On the Caltech data set, our approach outperforms all existing pedestrian detectors by a large margin.}
Spatial distribution of selected visual features is shown in Fig.~\ref{fig:filter} (left).
Each pixel is normalized such that that white pixels indicate most frequently chosen regions.
We observe that active regions focus mainly on head, shoulders and feet.
Similar observation has also been reported in \cite{Benenson2013Seeking}, in which
the authors apply a multi-scale model for pedestrian detection.
The training time of our approach is under $24$ hours on a parallelized
quad core Intel Xeon processor.

\begin{figure}[t]
        \centering
        \includegraphics[width=0.5\textwidth,clip]{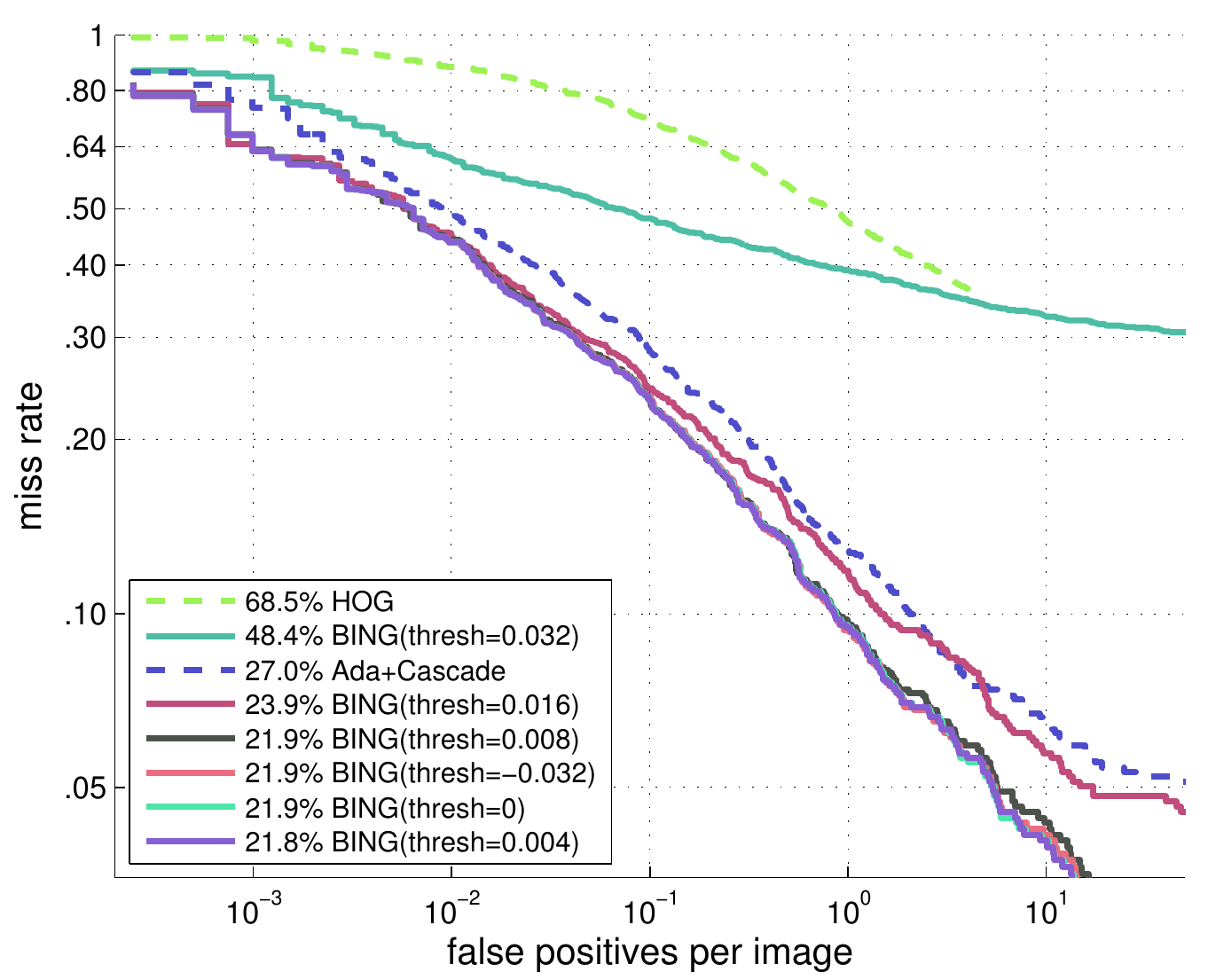}
    \caption{
    The change in the detection performance
    as we vary the threshold value of the BING detector
    (evaluated on the Caltech pedestrian test set).
    BING(thresh=$0$) represents the proposed two-stage pedestrian
    detector, in which the first stage is the BING classifier
    with the threshold value of zero and 
    and the second stage is the \algonametight detector described in
    Section~\ref{sec:exp_ped}.
    }
    \label{fig:bing}
\end{figure}

\paragraph{Region proposals generation}
In this section, we train a two-stage pedestrian detector
by placing the efficient BING classifier in the first stage and the
previously trained \algonametight pedestrian detector in the second stage.
To train the BING detector,
the resolution of the pedestrian model is set to $8 \times 16$ pixels.
The learned linear SVM model using BING features is shown in Fig.~\ref{fig:filter} (right).
We observe that most active pixels (white pixels) are near the human contour.
The SVM weights shown here are also similar to the learned SVM weights
of HOG (Fig.~6b in \cite{Dalal2005HOG}).
We compare the performance of our two-stage
pedestrian detector by varying the threshold value of the
BING detector in the first stage (varying the number of region
proposals being generated).
We plot ROC curves of our detector with different BING
threshold values in Fig.~\ref{fig:bing}.
From the figure, the performance starts to drop as we
increase the BING threshold value
(reducing the number of region proposals generated).
However we observe that setting the BING threshold value in
the range $[-0.004,0.008]$ results in similar pedestrian detection performance.
This clearly demonstrates that the BING detector can be applied to discard a large number
of background patches while retaining most pedestrian patches.
Table~\ref{tab:bing} compares the number of region proposals discarded in the first stage,
log-average miss rate and the average scanning time (excluding feature
extraction and post-processing computation, \eg, non-maximum suppression) of our two-stage detector by
varying the threshold value of the BING classifier
on the Caltech-USA test set ($640 \times 480$-pixel images).
From the table, setting the threshold value of the BING classifier
to be $0.008$ yields similar results to the original \algonametight detector
while reducing the window scanning time by half.
We observe a slight improvement in the log-average miss rate of $0.1\%$
when we set the BING threshold value to $0.004$.
We suspect that the BING detector might have discarded a
few difficult-to-classify background patches that the \algonametight detector fails to classify.

Next we compare the performance and evaluation time of our two-stage detector with
a soft cascade \cite{Bourdev05softcascade}.
For soft cascade, we train AdaBoost with a combination of
low-level visual features previously used.
All other experimental settings are kept the same (\eg,
a number of weak classifiers,
a number of bootstrapping iterations,
post-processing computation, \etc).
We heuristically set the soft cascade's rejection threshold at every node
to be $\{-160$,$-80$,$-40$,$-20$,$-10$,$-1\}$.
The performance and window scanning time of soft cascade with various rejection thresholds is shown in
Table~\ref{tab:softcascade}.
We observe that the cascaded classifier performs worse than our two-stage detector
(up to $5\%$ worse in terms of the log-average miss rate on the Caltech-USA benchmark).
Note that soft cascade (top row in Table~\ref{tab:softcascade}) has
a higher window scanning time than our two-stage approach
(top row in Table~\ref{tab:bing}).
The reason is that, for soft cascade, the partial sum of
weak classifiers' coefficients is repeatedly compared with the rejection threshold.
This additional comparison increases the window scanning time of soft cascade
when the rejection threshold is set to be small.

It is important to point out that our performance gain comes at the cost of
an increase in window scanning time.
For example, our detector achieves an average miss rate of $23.9\%$
with an average scan time of $2$ seconds per $640 \times 480$-pixel image while
soft cascade achieves an average miss rate of $27.1\%$ with an average scan
scan time of $0.3$ seconds per image.

\begin{table}[bt]
  \caption{Proportion of windows rejected by tuning the threshold of the BING classifier
  }
  \centering
  {
  \begin{tabular}{c|c|c|c}
  \hline
    BING & $\%$ windows & Log-avg.  & Avg. scan time \\
    threshold & discarded     & miss rate & per image (secs)      \\
  \hline
  \hline
   $-0.032$ & $0\%$ & $21.9\%$ & $5.8$ \\
   $0$ &     $13.4\%$ & $21.9\%$ & $5.1$ \\
   $0.004$ & $37.4\%$ & $21.8\%$ & $3.6$ \\
   $0.008$ & $49.4\%$ & $21.9\%$ & $2.8$ \\
   $0.016$ & $65.4\%$ & $23.9\%$ & $2.0$ \\
   $0.032$ & $86.6\%$ & $48.4\%$ & $0.8$ \\
  \hline
  \end{tabular}
  }
  \label{tab:bing}
\end{table}

\begin{table}[bt]
  \caption{Log-average miss rate and evaluation time of
           various AdaBoost based pedestrian detectors with
           different soft cascade's rejection thresholds
  }
  \centering
  {
  \begin{tabular}{c|c|c}
  \hline
    Soft cascade's & Log-avg.  & Avg. scan time \\
    rejection threshold & miss rate & per image (secs)      \\
  \hline
  \hline
   $-160$ &  $27.0\%$ & $6.90$ \\
   $-80$ &  $27.0\%$ & $3.68$ \\
   $-40$ &  $27.0\%$ & $1.62$ \\
   $-20$ &  $27.0\%$ & $0.71$ \\
   $-10$ &  $27.1\%$ & $0.31$ \\
   $-1$ &   $29.6\%$ & $0.02$ \\
  \hline
  \end{tabular}
  }
  \label{tab:softcascade}
\end{table}

\section{Conclusion}
\label{sec:con}
In this paper, we have proposed an approach to strengthen the effectiveness
of low-level visual features and formulated a new ensemble learning method
for object detection.
The proposed approach is combined with the efficient proposal generation,
which results in the effective classifier which optimizes the average miss rate performance measure.
Extensive experiments demonstrate the effectiveness of the
proposed approach on both synthetic data and visual detection tasks.
We plan to explore the possibility of applying the
proposed approach to the multiple scales detector of \cite{Benenson2012Pedestrian}
in order to improve the detection results of low
resolution pedestrian images.

\section*{Acknowledgements}

  This work was in part supported by the Data to
  Decisions Cooperative Research Centre.
  C. Shen's participation
  was in part supported by an Australian Research Council Future Fellowship.  
  C. Shen is
  the corresponding author.

\newpage
\onecolumn

\section*{Appendix}

\subsection{Convergence analysis of Algorithm 1  }

  In this Appendix, we provide a theoretical analysis of the convergence property for the structured ensemble learning
  in this paper.

  The main result is as follows.

  \begin{prop}
      At each iteration of Algorithm 1, the objective value decreases.
  \end{prop}

  {\bf Proof:}

  We assume that the current solution is a finite subset of weak learners and their
  corresponding coefficients are $ \bw $.
  If at the next iteration one more different weak learner is added into the current weak learner subset,
  and we re-solve the primal  optimization problem, and the corresponding $ \hat w$ is zero, then
  the objective value and the solution would be unchanged. If this happens, the current solution $ \bw $
  is already the optimal solution---one is not able to find another weak learner to decrease the objective value.

  Now if the corresponding $ \hat w$ is not zero, we have added one more free variable into the primal master problem,
  and re-solving it must reduce the objective value.

  With the next proposition, we show that the convergence of Algorithm 1
  is guaranteed.
  \begin{prop}
  The decrease of objective value between iterations $ t - 1 $ and $ t $ is not less than
\[
 \Bigl[  \phizeta ( \bh_{t:}, \bYast ) - \phizeta
        ( \bh_{t:}, \bY_{[t]}^\star)
      \Bigr]^2.
\]
    Here,
\begin{equation*}
\bY_{[t]}^\star  =  \argmax_{  \bY  }
  \Bigl\{
  \deltapauc (\bY, \bYast)  +   \bw_{[t]}  ^ \T  \phizeta (\bH, \bY)
\Bigr\},
\end{equation*}
and the subscript $[t]$ denotes the index at iteration $t$.
  \end{prop}

  {\bf Proof:}

Recall that the optimization problem we want to solve is:
\begin{align}
\label{EQ:hinge1a2}
    \min_{ \bw , \xi }   \quad
    &
    \frac{1}{2} \| \bw  \|_{2}^{2} + \nu \, \xi    \\ \notag
    \st \; &
    \bw^\T ( \phizeta (\bH, \bYast) - \phizeta (\bH, \bY) )
    \geq \deltapauc (\bY, \bYast) - \xi,
\end{align}
$\forall \bY \in \sY_{m,\jbeta}$ and $\xi \geq 0$.

Here
$\bH = (\bHp, \bHm)$ is the projected output
for positive and negative training samples.
$\phizeta (\bH, \bY) = [\phizeta (\bh_{1:}, \bY), \cdots, \phizeta (\bh_{k:}, \bY)]$
where $\phizeta (\bh_{t:}, \bY): (\Real^{m} \times \Real^{n}) \times \sY_{m,\jbeta} \rightarrow \Real$
and it is defined as,
\begin{align}
    \label{EQ:featmap21}
    \phizeta (\bh_{t:}, \bY) = \frac{1}{c} \;
            \sum_{i=1}^m  \sum_{j=1}^{\jbeta} (1 - \yij)
            \bigl( \hbar_{t} (\bxip) - \hbar_{t} (\bxkjm) \bigr),
\end{align}
where $\{ \bxkjm \}_{j=1}^{\jbeta}$ is any given subsets of negative instances
and $\boldsymbol{k} = [k_1,\ldots ,k_{\jbeta}]$ is a vector indicating which elements of $\mS_{-}$ are included.

At iteration $t$ in Algorithm 1,
  the primal objective in Equation \eqref{EQ:hinge1a2} can be reformulated into:
  \begin{align}
  F ( \bw_{[t]} )
  &=
  \frac{1}{2} \sum_{ \tau = 1 } ^ t w_{[t],\tau}^2  +  \nu      \cdot
  \max_{  \bY \in \sY_{m,\jbeta}       }
  \Bigl\{
    \deltapauc (\bY, \bYast) -   \bw_{[t]}^\T \bigl[ \phizeta (\bH, \bYast) - \phizeta (\bH, \bY) \bigr]
\Bigr\}
\notag
\\
&=  \frac{1}{2} \sum_{ \tau = 1 } ^ t w_{[t],\tau}^2  +  \nu      \cdot
\max_{  \bY  }
  \Bigl\{
  \deltapauc (\bY, \bYast)  +   \bw_{[t]}  ^\T  \phizeta (\bH, \bY)
\Bigr\}
- \nu \bw_{[t]}^\T   \phizeta (\bH, \bYast),
\end{align}
where $\bw_{[t]}$ denotes the optimal solution at iteration $t$ and
$\bw_{[t]} = \bigl[w_{[t],1}, w_{[t],2}, \cdots, w_{[t],t} \bigr] ^\T $.

Let us define
\begin{equation}
\bY_{[t]}^\star  =  \argmax_{  \bY  }
  \Bigl\{
  \deltapauc (\bY, \bYast)  +   \bw_{[t]}  ^ \T  \phizeta (\bH, \bY)
\Bigr\}.
\label{def0}
\end{equation}
Now the objective function at iteration $ t $ is
\begin{equation}
  F ( \bw_{[t]},  \bY_{[t]}^\star   )
  =
  \frac{1}{2} \sum_{ \tau = 1 } ^ t w_{[t],\tau}^2  +  \nu      \cdot
    \deltapauc ( \bY_{[t]}^\star, \bYast) -   \bw_{[t]}^\T \bigl[ \phizeta (\bH, \bYast)
    - \phizeta (\bH, \bY_{[t]}^\star) \bigr].
    \label{def}
  \end{equation}
We know that $  \bY_{[t]}^\star   $ is a sub-optimal maximization solution for iteration $ (t - 1 )$.
Therefore the following inequality must hold:
\begin{equation}
F ( \bw_{[t-1]} ) - F ( \bw_{[t]} )
=  F ( \bw_{[t-1]},  \bY_{[t-1]}^\star    ) - F ( \bw_{[t]},  \bY_{[t]}^\star  )
\geq  F ( \bw_{[t-1]},  \bY_{[t]}^\star    ) - F ( \bw_{[t]},  \bY_{[t]}^\star  ).
\label{eq:ineq1}
\end{equation}

Now with an arbitrary value $ \omega $, we know that
\[
  \tilde \bw_{[ t ]} = \begin{bmatrix}
                        \bw_{[ t -1 ]}    \\
                      \omega
                      \end{bmatrix}
\]
is a sub-optimal solution for iteration $ t $.
Here $ \bw_{[ t -1 ]}  $ is the optimal solution for iteration $ t - 1 $.
The inequality in \eqref{eq:ineq1} continues as
\begin{equation}
  F ( \bw_{[t-1]} ) - F ( \bw_{[t]} ) = \dots \geq \dots
  \geq  F ( \bw_{[t-1]},   \bY_{[t]}^\star  ) - F ( \tilde \bw_{[t]} ,  \bY_{[t]}^\star  ).
  \label{eq:ineq2}
\end{equation}
With the above definition \eqref{def},
We can greatly simplify \eqref{eq:ineq2}, which is
\[
  {\rm r.h.s.~of~\eqref{eq:ineq2}}  =
  - \tfrac{1}{2} \omega^2   +\omega  \Bigl[  \phizeta ( \bh_{t:}, \bYast ) - \phizeta   ( \bh_{t:}, \bY_{[t]}^\star)
  \Bigr].
\]

In summary,
the objective decrease is lower bounded as:
\begin{align*}
F ( \bw_{[t-1]} ) - F ( \bw_{[t]} )
& \geq
 \max_\omega
 \Bigl\{
      - \tfrac{1}{2} \omega^2   +\omega  \Bigl[  \phizeta ( \bh_{t:}, \bYast ) - \phizeta
        ( \bh_{t:}, \bY_{[t]}^\star)
        \Bigr]
 \Bigr\}
 \\
&=    \Bigl[  \phizeta ( \bh_{t:}, \bYast ) - \phizeta
        ( \bh_{t:}, \bY_{[t]}^\star)
        \Bigr]^2
\end{align*}
where  $\bY_{[t]}^\star$   is calculated by \eqref{def0}.

  Note that, since the structured ensemble learning method in \cite{Paul2013Efficient},
  the analysis here can be easily adapted
  so that it applies to \cite{Paul2013Efficient}.

\twocolumn

{
\bibliographystyle{IEEEtran}
\bibliography{draft}
}

\begin{IEEEbiographynophoto}{Sakrapee Paisitkriangkrai}
is a postdoctoral researcher at The Australian Centre for Visual Technologies, The University of Adelaide. 
He  received his Bachelor degree in computer engineering, 
the Master degree in biomedical engineering, and the PhD degree 
from the University of New South Wales, Sydney, Australia, in 2003 and 2010, respectively.
His research interests include pattern recognition, image processing, and machine learning.
\end{IEEEbiographynophoto}

\begin{IEEEbiographynophoto}{Chunhua Shen}
is a Professor at School of Computer Science, The
University of Adelaide.  His research interests are in the
intersection of computer vision and statistical machine learning.

He studied at Nanjing University, at Australian National University,
and received his PhD degree from The University of Adelaide. In 2012, 
he was awarded the Australian Research Council Future Fellowship.
\end{IEEEbiographynophoto}

\begin{IEEEbiographynophoto}{Anton van den Hengel}
is the Founding Director of The Australian
Centre for Visual Technologies, at the University of Adelaide, focusing
on innovation in the production and analysis of
visual digital media.

He received the Bachelor of
mathematical science degree, Bachelor of laws
degree, Master's degree in computer science,
and the PhD degree in computer vision from The
University of Adelaide in 1991, 1993, 1994, and
2000, respectively.
\end{IEEEbiographynophoto}

\end{document}